%% file: main.tex
\documentclass[runningheads]{llncs}
\usepackage{graphicx}
\usepackage{comment}
\usepackage{amsmath,amssymb} 
\usepackage{color}
\usepackage{algorithm, algorithmic}

\usepackage{siunitx}
\DeclareSIUnit{\nothing}{\relax}
\usepackage{gensymb}

\usepackage{xfrac}

\usepackage{amsmath}

\usepackage{tabularx}

\usepackage{multirow}
\usepackage{booktabs}  
\usepackage{colortbl}
\usepackage{xcolor,colortbl}

\usepackage{caption}
\usepackage{subcaption}
\usepackage{verbatim}
\usepackage{xspace}
\usepackage{wrapfig}
\usepackage[colorlinks=true,linkcolor=blue,urlcolor=blue]{hyperref}

\makeatletter
\DeclareRobustCommand\onedot{\futurelet\@let@token\@onedot}
\def\@onedot{\ifx\@let@token.\else.\null\fi\xspace}

\makeatother
 
\newcommand{\name}{{BiT}}
\newcommand{\hyper}{{BiT-HyperRule}}
\newcommand{\imagenet}{{ILSVRC-2012}}

\DeclareSIUnit\px{px}

\newcolumntype{C}{>{\centering\arraybackslash}X}

\begin{document}
\pagestyle{headings}
\mainmatter
\def\ECCVSubNumber{3665}  

\title{Big Transfer (BiT): \\ General Visual Representation Learning}

\titlerunning{Big Transfer (BiT): General Visual Representation Learning}
\author{Alexander Kolesnikov\makebox[0pt]{\hspace{5pt}\thanks{Equal contribution}} \and
Lucas Beyer\makebox[0pt]{\hspace{5pt}$^\star$} \and
Xiaohua Zhai\makebox[0pt]{\hspace{5pt}$^\star$} \and \\
Joan Puigcerver\makebox[0pt]{\hspace{5pt}} \and
Jessica Yung \and
Sylvain Gelly \and
Neil Houlsby}
\authorrunning{Kolesnikov\makebox[0pt]{\hspace{5pt}$^\star$}, Beyer\makebox[0pt]{\hspace{5pt}$^\star$}, Zhai\makebox[0pt]{\hspace{5pt}$^\star$}, Puigcerver, Yung, Gelly, Houlsby}

\institute{Google Research, Brain Team\\  Z\"urich, Switzerland \\
\email{\{akolesnikov,lbeyer,xzhai\}@google.com} \\
\email{\{jpuigcerver,jessicayung,sylvaingelly,neilhoulsby\}@google.com}
}

\maketitle              
\input{content}

\bibliographystyle{splncs04}
\bibliography{main}

\clearpage

\input{appendix}

\end{document}

%% file: content.tex
\begin{abstract}
Transfer of pre-trained representations improves sample efficiency and simplifies hyperparameter tuning when training deep neural networks for vision. 
We revisit the paradigm of pre-training on large supervised datasets and fine-tuning the model on a target task. 
We scale up pre-training, and propose a simple recipe that we call Big Transfer (\name{}).
By combining a few carefully selected components, and transferring using a simple heuristic, we achieve strong performance on over 20 datasets.
\name{} performs well across a surprisingly wide range of data regimes --- from 1 example per class to 1\,M total examples. 
\name{} achieves 87.5\% top-1 accuracy on \imagenet{}, 99.4\% on CIFAR-10, and 76.3\% on the 19 task Visual Task Adaptation Benchmark (VTAB).
On small datasets, \name{} attains 76.8\% on \imagenet{} with 10 examples per class, and 97.0\% on CIFAR-10 with 10 examples per class.
We conduct detailed analysis of the main components that lead to high transfer performance.
\end{abstract}

\section{Introduction}
Strong performance using deep learning usually requires a large amount of task-specific data and compute.
These per-task requirements can make new tasks prohibitively expensive.
Transfer learning offers a solution: 
task-specific data and compute are replaced with a pre-training phase.
A network is trained once on a large, generic dataset, and its weights are then used to initialize subsequent tasks which can be solved with fewer data points, and less compute~\cite{pan2009survey,raghu2019,he2019rethinking}.

\begin{figure}[t]
  \centering
  \includegraphics[width=1.0\textwidth]{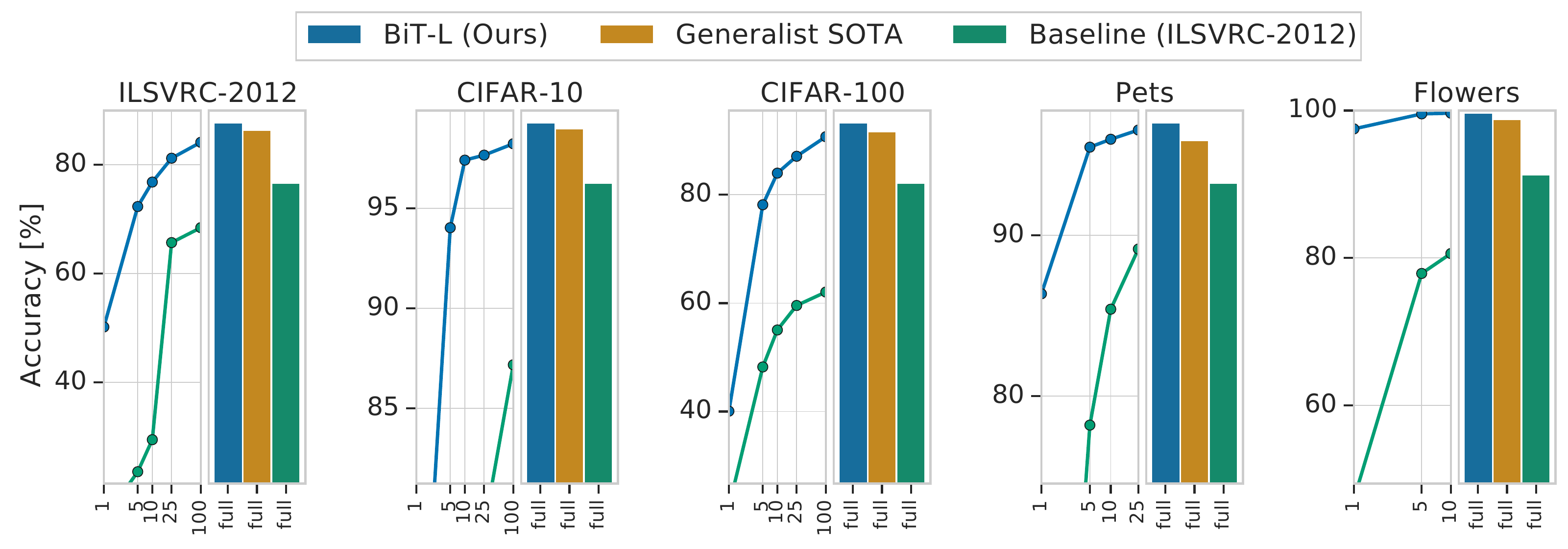}
  \caption{%
Transfer performance of our pre-trained model, \name{}-L, the previous state-of-the-art (SOTA), and a ResNet-50 baseline pre-trained on \imagenet{} to downstream tasks.
Here we consider only methods that are pre-trained independently of the final task (generalist representations), like \name{}.
The bars show the accuracy when fine-tuning on the full downstream dataset.
The curve on the left-hand side of each plot shows that \name{}-L performs well even when transferred using only few images (1 to 100) per class.
}\label{fig:teaser_plot}
\vspace{-3mm}
\end{figure}

We revisit a simple paradigm:
pre-train on a large supervised source dataset, and fine-tune the weights on the target task.
Numerous improvements to deep network training have recently been introduced, e.g.~\cite{tan2019efficientnet,xie2017aggregated,kingma2014adam,loshchilov2016sgdr,izmailov2018averaging,fast_swa,yun2019cutmix,mixup,szegedy2016rethinking,wu2018group}.
We aim not to introduce a new component or complexity, but to provide a recipe that uses the minimal number of tricks yet attains excellent performance on many tasks.
We call this recipe ``Big Transfer'' (\name{}).

We train networks on three different scales of datasets.
The largest, \name{}-L is trained on the JFT-300M dataset~\cite{sun2017revisiting}, which contains 300\,M noisily labelled images.
We transfer \name{} to many diverse tasks; with training set sizes ranging from 1 example per class to 1M total examples.
These tasks include ImageNet's ILSVRC-2012~\cite{deng2009imagenet}, CIFAR-10/100~\cite{cifar10}, Oxford-IIIT Pet~\cite{parkhi12a}, Oxford Flowers-102~\cite{Nilsback08} (including few-shot variants), and the 1000-sample VTAB-1k benchmark~\cite{zhai2019visual}, which consists of 19 diverse datasets.
\name{}-L attains state-of-the-art performance on many of these tasks, and is surprisingly effective when very little downstream data is available (Figure~\ref{fig:teaser_plot}).
We also train \name{}-M on the public ImageNet-21k dataset, and attain marked improvements over the popular \imagenet{} pre-training.

Importantly, \name{} only needs to be pre-trained once and subsequent fine-tuning to downstream tasks is cheap.
By contrast, other state-of-the-art methods require extensive training on support data conditioned on the task at hand \cite{dat,noisystudent,yalniz2019billion}.
Not only does \name{} require a short fine-tuning protocol for each new task, but \name{} also does not require extensive hyperparameter tuning on new tasks.
Instead, we present a heuristic for setting the hyperparameters for transfer, which works well on our diverse evaluation suite.

We highlight the most important components that make Big Transfer effective, and provide insight into the interplay between scale, architecture, and training hyperparameters.
For practitioners, we will release the performant \name-M model trained on ImageNet-21k.

\section{Big Transfer}
\label{sec:methods}

We review the components that we found necessary to build an effective network for transfer.
\emph{Upstream} components are those used during pre-training, and \emph{downstream} are those used during fine-tuning to a new task.

\subsection{Upstream Pre-Training}
\label{sec:methods_upstream}

The first component is scale.
It is well-known in deep learning that larger networks perform better on their respective tasks~\cite{he2016deep,simonyan2014very}.
Further, it is recognized that larger datasets require larger architectures to realize benefits, and vice versa~\cite{kaplan2020scaling,rosenfeld2019constructive}.
We study the effectiveness of scale (during pre-training) in the context of transfer learning, including transfer to tasks with very few datapoints.
We investigate the interplay between computational budget (training time), architecture size, and dataset size.
For this, we train three \name{} models on three large datasets: 
\imagenet{}~\cite{imagenet} which contains 1.3M images (\name{}-S), ImageNet-21k~\cite{deng2009imagenet} which contains 14M images (\name{}-M),
and JFT~\cite{sun2017revisiting} which contains 300M images (\name{}-L).

The second component is Group Normalization (GN)~\cite{wu2018group} and Weight Standardization (WS)~\cite{lin2014microsoft}.
Batch Normalization (BN)~\cite{ioffe2015batch} is used in most state-of-the-art vision models to stabilize training.
However, we find that BN is detrimental to Big Transfer for two reasons.
First, when training large models with small per-device batches, BN performs poorly or incurs inter-device synchronization cost.
Second, due to the requirement to update running statistics, BN is detrimental for transfer.
GN, when combined with WS, has been shown to improve performance on small-batch training for ImageNet and COCO~\cite{lin2014microsoft}.
Here, we show that the combination of GN and WS is useful for training with large batch sizes, and has a significant impact on transfer learning.

\subsection{Transfer to Downstream Tasks}
\label{sec:methods_downstream}

We propose a cheap fine-tuning protocol that applies to many diverse downstream tasks.
Importantly, we avoid expensive hyperparameter search for every new task and dataset size; we try only one hyperparameter per task.
We use a heuristic rule---which we call \hyper{}---to select the most important hyperparameters for tuning as a simple function of the task's intrinsic image resolution and number of datapoints.
We found it important to set the following hyperparameters per-task: training schedule length, resolution, and whether to use MixUp regularization~\cite{mixup}.
We use \hyper{} for over 20 tasks in this paper, with training sets ranging from 1 example per class to over 1M total examples.
The exact settings for \hyper{} are presented in Section~\ref{sec:hyper}.

During fine-tuning, we use the following standard data pre-processing:
we resize the image to a square, crop out a smaller random square, and randomly horizontally flip the image at training time.
At test time, we only resize the image to a fixed size.
In some tasks horizontal flipping or cropping destroys the label semantics, making the task impossible. An example is if the label requires predicting object orientation or coordinates in pixel space.
In these cases we omit flipping or cropping when appropriate.

Recent work has shown that existing augmentation methods introduce inconsistency between training and test resolutions for CNNs~\cite{fixres}.
Therefore, it is common to scale up the resolution by a small factor at test time.
As an alternative, one can add a step at which the trained model is fine-tuned to the test resolution~\cite{fixres}.
The latter is well-suited for transfer learning; we include the resolution change during our fine-tuning step.

We found that MixUp~\cite{mixup} is not useful for pre-training \name{}, likely due to the abundance of data.
However, it is sometimes useful for transfer.
Interestingly, it is most useful for mid-sized datasets, and not for few-shot transfer, see Section~\ref{sec:hyper} for where we apply MixUp.

Surprisingly, we do not use any of the following forms of regularization during downstream tuning: weight decay to zero, weight decay to initial parameters~\cite{li2018explicit}, or dropout. Despite the fact that the network is very large---\name{} has 928 million parameters---the performance is surprisingly good without these techniques and their respective hyperparameters, even when transferring to very small datasets.
We find that setting an appropriate schedule length, i.e.\ training longer for larger datasets, provides sufficient regularization.

\section{Experiments}

We train three upstream models using three datasets at different scales: \name{}-S, \name{}-M, and \name{}-L.
We evaluate these models on many downstream tasks and attain very strong performance on high and low data regimes.

\subsection{Data for Upstream Training}
\name{}-S is trained on the \imagenet{} variant of ImageNet, which contains $1.28$ million images and $1000$ classes.
Each image has a single label. 
\name{}-M is trained on the full ImageNet-21k dataset~\cite{deng2009imagenet}, a public dataset containing 14.2 million images and 21k classes organized by the WordNet hierarchy.
Images may contain multiple labels.
\name{}-L is trained on the JFT-300M dataset~\cite{sun2017revisiting,dat,noisystudent}.
This dataset is a newer version of that used in~\cite{hinton2015distilling,chollet2017xception}.
JFT-300M consists of around 300 million images with 1.26 labels per image on average.
The labels are organized into a hierarchy of $18\,291$ classes.
Annotation is performed using an automatic pipeline, and are therefore imperfect; approximately 20\% of the labels are noisy.
We remove all images present in downstream test sets from JFT-300M. See Supplementary Material section~\ref{sec:dedup} for details.
Note: the ``-S/M/L'' suffix refers to the pre-training datasets size and schedule, not architecture. We train \name{} with several architecture sizes, the default (largest) being ResNet152x4.

\subsection{Downstream Tasks}

We evaluate \name{} on long-standing benchmarks: ILSVRC-2012~\cite{deng2009imagenet}, CIFAR-10/100 \cite{cifar10}, Oxford-IIIT Pet~\cite{parkhi12a} and Oxford Flowers-102~\cite{Nilsback08}.
These datasets differ in the total number of images, input resolution and nature of their categories, from general object categories in ImageNet and CIFAR to fine-grained ones in Pets and Flowers. 
We fine-tune \name{} on the official training split and report results on the official \emph{test} split if publicly available. Otherwise, we use the \emph{val} split.

To further assess the generality of representations learned by \name{}, we evaluate on the Visual Task Adaptation Benchmark (VTAB)~\cite{zhai2019visual}.
VTAB consists of 19 diverse visual tasks, each of which has 1000 training samples (VTAB-1k variant). The tasks are organized into three groups: \emph{natural}, \emph{specialized} and \emph{structured}. The VTAB-1k score is top-1 recognition performance averaged over these 19 tasks. The \emph{natural} group of tasks contains classical datasets of natural images captured using standard cameras. The \emph{specialized} group also contains images captured in the real world, but through specialist equipment, such as satellite or medical images. Finally, the \emph{structured} tasks assess understanding of the the structure of a scene, and are mostly generated from synthetic environments. Example structured tasks include object counting and 3D depth estimation.

\begin{table}[t]
  \setlength{\tabcolsep}{0pt}
  \setlength{\extrarowheight}{5pt}
  \renewcommand{\arraystretch}{0.75}
  \centering
  \caption{Top-1 accuracy for \name{}-L on many datasets using a single model and single hyperparameter setting per task (\hyper{}). 
  The entries show median $\pm$ standard deviation across 3 fine-tuning runs. Specialist models are those that condition pre-training on each task, while generalist models, including \name{}, perform task-independent pre-training. ($^\star$Concurrent work.)}\label{tbl:main}
  \begin{tabularx}{\linewidth}{p{3.0cm}p{0.2cm}Cp{0.2cm}Cp{1cm}C}
    \cmidrule[1pt]{1-5} \cmidrule[1pt]{7-7}
     && \name{}-L && Generalist SOTA && Specialist SOTA \\
    \cmidrule[0.5pt]{1-5} \cmidrule[0.5pt]{7-7}

    \imagenet{} && \textbf{87.54 $\pm$ 0.02} && 86.4~\cite{fixres} && 88.4~\cite{noisystudent}$^\star$ \\
    CIFAR-10 && \textbf{99.37 $\pm$ 0.06} && 99.0~\cite{gpipe} && - \\
    CIFAR-100 && \textbf{93.51 $\pm$ 0.08} && 91.7~\cite{tan2019efficientnet} && - \\

    Pets && \textbf{96.62 $\pm$ 0.23} && 95.9~\cite{gpipe} && 97.1~\cite{dat} \\
    Flowers && \textbf{99.63 $\pm$ 0.03} && 98.8~\cite{tan2019efficientnet} && 97.7~\cite{dat} \\

    VTAB (19 tasks) && \textbf{76.29 $\pm$ 1.70} && 70.5~\cite{vivi} && - \\
    \cmidrule[1pt]{1-5} \cmidrule[1pt]{7-7}
  \end{tabularx}
\vspace{-4mm}
\end{table}

\subsection{Hyperparameter Details}\label{sec:hyper}

\subsubsection{Upstream Pre-Training}
All of our BiT models use a vanilla ResNet-v2 architecture~\cite{he2016identity}, except that we replace all Batch Normalization~\cite{ioffe2015batch} layers with Group Normalization~\cite{wu2018group} and use Weight Standardization~\cite{qiao2019weight} in all convolutional layers.
See Section~\ref{sec:gn} for analysis.
We train ResNet-152 architectures in all datasets, with every hidden layer widened by a factor of four (ResNet152x4).
We study different model sizes and the coupling with dataset size in Section~\ref{sec:architecture}.

We train all of our models upstream using SGD with momentum.
We use an initial learning rate of $0.03$, and momentum $0.9$.
During image preprocessing stage we use image cropping technique from~\cite{szegedy2015going} and random horizontal mirroring followed by $224 \times 224$ image resize.
We train both \name{}-S and \name{}-M for 90 epochs and decay the learning rate by a factor of 10 at 30, 60 and 80 epochs.
For \name{}-L, we train for 40 epochs and decay the learning rate after 10, 23, 30 and 37 epochs.
We use a global batch size of 4096 and train on a Cloud TPUv3-512~\cite{jouppi2017datacenter}, resulting in 8 images per chip.
We use linear learning rate warm-up for 5000 optimization steps and multiply the learning rate by $\frac{\mbox{batch size}}{256}$ following~\cite{goyal2017accurate}. 
During pre-training we use a weight decay of $0.0001$, but as discussed in Section~\ref{sec:methods}, we do not use any weight decay during transfer.

\subsubsection{Downstream Fine-Tuning}
\label{sec:bit-hp}
To attain a low per-task adaptation cost, we do not perform any hyperparameter sweeps downstream.
Instead, we present \hyper{}, a heuristic to determine all hyperparameters for fine-tuning.
Most hyperparameters are fixed across all datasets, but schedule, resolution, and usage of MixUp depend on the task’s image resolution and training set size.

For all tasks, we use SGD with an initial learning rate of 0.003, momentum 0.9, and batch size 512.
We resize input images with area smaller than $96\times96$ pixels to $160\times160$ pixels, and then take a random crop of $128\times128$ pixels.
We resize larger images to $448\times448$ and take a $384\times384$-sized crop.\footnote{For our largest R152x4, we increase resolution to $512\times512$ and crop to $480\times480$.}
We apply random crops and horizontal flips for all tasks, except those for which cropping or flipping destroys the label semantics, see Supplementary section~\ref{sec:flip-and-crop-details} for details.

For schedule length, we define three scale regimes based on the number of examples:
we call \emph{small} tasks those with fewer than 20\,k labeled examples, 
\emph{medium} those with fewer than 500\,k, 
and any larger dataset is a \emph{large} task.
We fine-tune \name{} for $500$ steps on small tasks, 
for 10k steps on medium tasks, 
and for 20k steps on large tasks.
During fine-tuning, we decay the learning rate by a factor of 10 at 30\%, 60\% and 90\% of the training steps.
Finally, we use MixUp~\cite{mixup}, with $\alpha = 0.1$, for medium and large tasks.
See Supplementary section~\ref{sec:tuningtransfer} for details.

\begin{table}[t]
  \setlength{\tabcolsep}{0pt}
  \setlength{\extrarowheight}{5pt}
  \renewcommand{\arraystretch}{0.75}
  \centering
  \caption{Improvement in accuracy when pre-training on the public ImageNet-21k dataset over the ``standard'' \imagenet{}.
  Both models are ResNet152x4.}\label{tbl:improvements}
  \begin{tabularx}{\linewidth}{p{2.7cm}p{0.17cm}Cp{0.17cm}Cp{0.17cm}Cp{0.17cm}Cp{0.17cm}Cp{0.17cm}C}
    \toprule[1pt]
     && \imagenet{} && CIFAR-10 && CIFAR-100 && Pets && Flowers && VTAB-1k (19 tasks)\\
    \midrule
    \name{}-S {\tiny (\imagenet{})}  && 81.30 && 97.51 && 86.21 && 93.97 && 89.89 && 66.87 \\
    \name{}-M {\tiny (ImageNet-21k)} && 85.39 && 98.91 && 92.17 && 94.46 && 99.30 && 70.64 \\
    \arrayrulecolor{lightgray}\midrule[0.25pt]\arrayrulecolor{black}
    Improvement                      && +4.09 && +1.40 && +5.96 && +0.49 && +9.41 && +3.77 \\
    \bottomrule
  \end{tabularx}
\vspace{-2mm}
\end{table}

\subsection{Standard Computer Vision Benchmarks}

We evaluate \name{}-L on standard benchmarks and compare its performance to the current state-of-the-art results (Table~\ref{tbl:main}).
We separate models that perform task-independent pre-training (``general'' representations), from those that perform task-dependent auxiliary training (``specialist'' representations).
The specialist methods condition on a particular task, for example \imagenet{}, then train using a large support dataset, such as JFT-300M~\cite{dat} or Instagram-1B~\cite{yalniz2019billion}.
See discussion in Section~\ref{sec:related}.
Specialist representations are highly effective, but require a large training cost \emph{per task}.
By contrast, generalized representations require large-scale training \emph{only once}, followed by a cheap adaptation phase.

\name{}-L outperforms previously reported generalist SOTA models as well as, in many cases, the SOTA specialist models.
Inspired by strong results of \name{}-L trained on JFT-300M, we also train models on the public ImageNet-21k dataset.
This dataset is more than 10 times bigger than \imagenet{}, but it is mostly overlooked by the research community.
In Table~\ref{tbl:improvements} we demonstrate that \name{}-M trained on ImageNet-21k leads to substantially improved visual representations compared to the same model trained on \imagenet{} (\name{}-S), as measured by all our benchmarks.
In Section~\ref{sec:schedule}, we discuss pitfalls that may have hindered wide adoption of ImageNet-21k as a dataset model for pre-training and highlight crucial components of \name{} that enabled success on this large dataset.   

For completeness, we also report top-5 accuracy on \imagenet{} with median $\pm$ standard deviation format across 3 runs: 98.46\%~$\pm$~0.02\% for \name{}-L, 97.69\%~$\pm$~0.02\% for \name{}-M and 95.65\%~$\pm$~0.03\% for \name{}-S.

\begin{figure}[t]
\centering
\includegraphics[width=\textwidth]{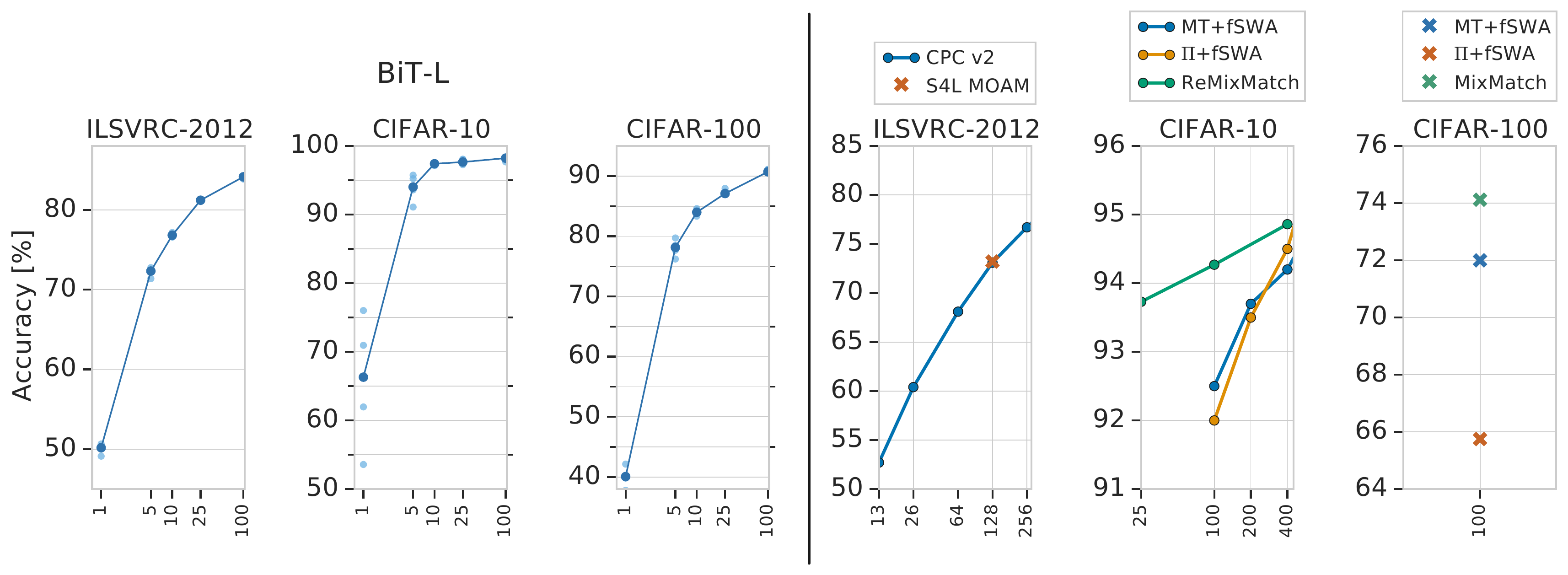}
\caption{
Experiments in the low data regime.
\textbf{Left:} Transfer performance of \name{}-L.
Each point represents the result after training on a balanced random subsample of the dataset (5 subsamples per dataset).
The median across runs is highlighted by the curves.
The variance across data samples is usually low, with the exception of 1-shot CIFAR-10, which contains only 10 images.
\textbf{Right:}
We summarize the state-of-the-art in semi-supervised learning as reference points.
Note that a direct comparison is not meaningful; 
unlike \name{}, semi-supervised methods have access to extra unlabelled data from the training distribution, but they do not make use of out-of-distribution labeled data.
}
\label{fig:low_data_plot}
\end{figure}

\subsection{Tasks with Few Datapoints}
We study the number of \emph{downstream} labeled samples required to transfer \name{}-L successfully.
We transfer \name{}-L using subsets of \imagenet{}, CIFAR-10, and CIFAR-100, down to 1 example per class.
We also evaluate on a broader suite of 19 VTAB-1k tasks, each of which has 1000 training examples.

Figure~\ref{fig:low_data_plot} (left half) shows \name{}-L using few-shots on \imagenet{}, CIFAR-10, and CIFAR-100. 
We run multiple random subsamples, and plot every trial.
Surprisingly, even with very few samples per class, \name{}-L demonstrates strong performance and quickly approaches performance of the full-data regime.
In particular, with just 5 labeled samples per class it achieves top-1 accuracy of 72.0\% on \imagenet{} and with 100 samples the top-1 accuracy goes to 84.1\%.
On CIFAR-100, we achieve 82.6\% with just 10 samples per class.

Semi-supervised learning also tackles learning with few labels.
However, such approaches are not directly comparable to \name{}.
\name{} uses extra labelled out-of-domain data, whereas semi-supervised learning uses extra unlabelled in-domain data.
Nevertheless, it is interesting to observe the relative benefits of transfer from out-of-domain labelled data versus in-domain semi-supervised data.
In Figure~\ref{fig:low_data_plot} we show state-of-the-art results from the semi-supervised learning. 

\begin{figure}[t]
\centering
\includegraphics[width=1.0\textwidth]{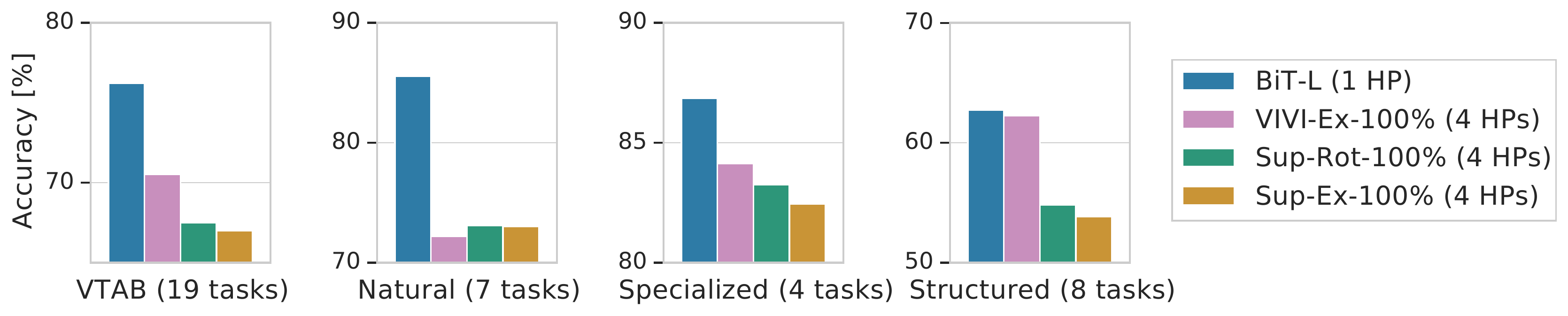}
\caption{Results on VTAB (19 tasks) with 1000 examples/task, and the current SOTA.
It compares methods that sweep few hyperparameters per task: either four hyperparameters in previous work (``4 HPs'') or the single \hyper{}.}\label{fig:vtab_1k_bar}\vspace{-1mm}
\end{figure}

Figure~\ref{fig:vtab_1k_bar} shows the performance of \name{}-L on the 19 VTAB-1k tasks.
\name{}-L with \hyper{} substantially outperforms the previously reported state-of-the-art.
When looking into performance of VTAB-1k task subsets, \name{} is the best on \emph{natural}, \emph{specialized} and \emph{structured} tasks.
The recently-proposed VIVI-Ex-100\%~\cite{vivi} model that employs video data during upstream pre-training shows very similar performance on the \emph{structured} tasks.

We investigate heavy per-task hyperparameter tuning in Supplementary Material section~\ref{sec:tuningtransfer} and conclude that this further improves performance.

\subsection{ObjectNet: Recognition on a ``Real-World'' Test Set}

\begin{wrapfigure}{r}{0.44\textwidth}
  \centering
  \vspace{-25pt}
  \includegraphics[width=0.44\textwidth]{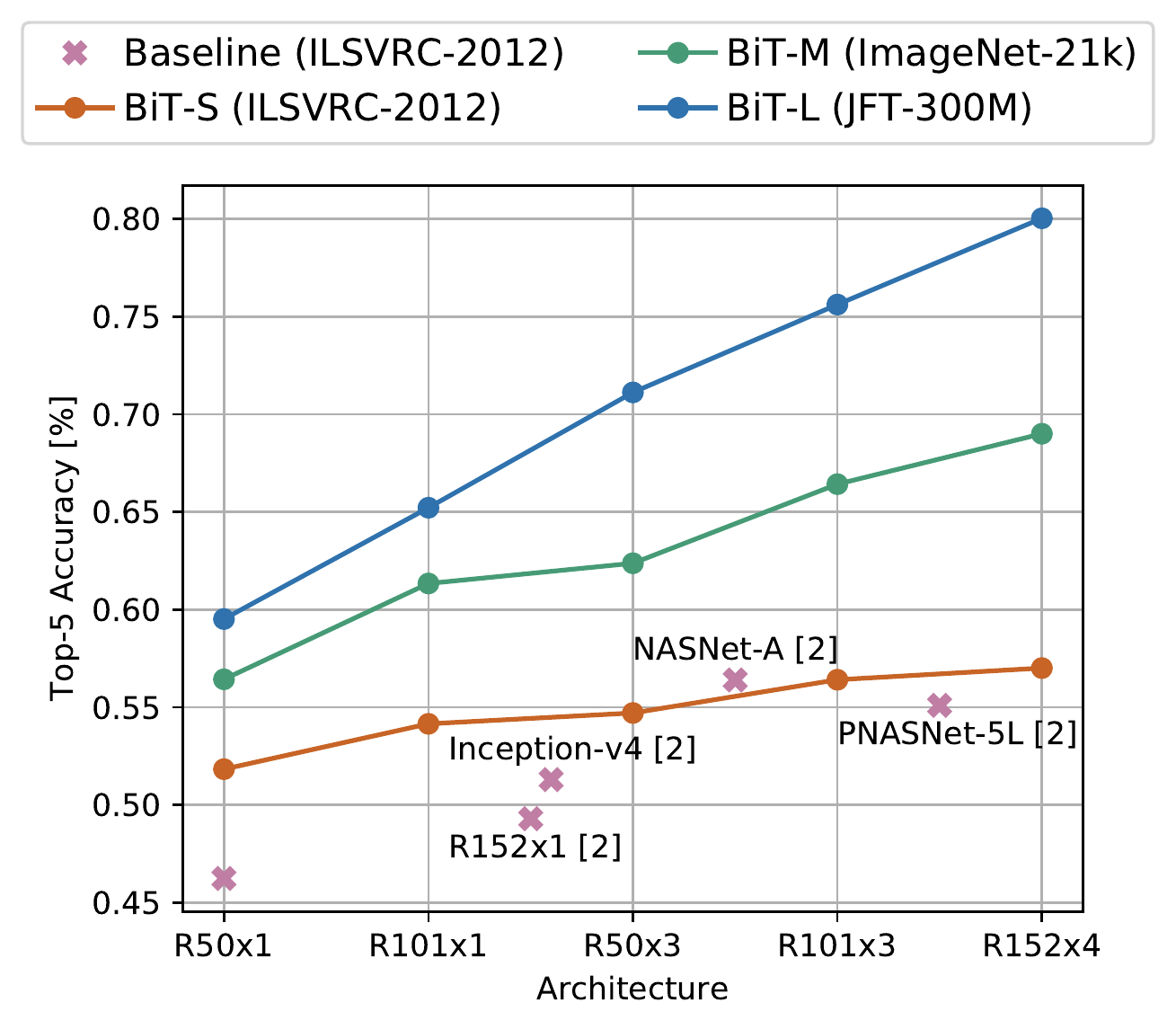}
  \caption{Accuracy of BiT models along with baselines on ObjectNet. R is short for ResNet in x-axis.}
\vspace{-20pt}
\label{fig:objectnet_top5}
\end{wrapfigure}

We evaluate \name{} on the new test-only ObjectNet dataset~\cite{barbu2019objectnet}.
Importantly, this dataset closely resembles real-life scenarios, where object categories may appear in non-canonical context, viewpoint, rotation, etc.
There are 313 object classes in total, with 113 overlapping with \imagenet{}. 
We follow the literature~\cite{barbu2019objectnet,borji2020reobjectnet}
and evaluate our models on those 113 classes. 

Figure~\ref{fig:objectnet_top5} shows that larger architectures and pre-training on more data results in higher accuracies.
Crucially, our results highlight that scaling both is essential for achieving unprecedented  top-5 accuracy of 80.0\%, an almost 25\% absolute improvement over the previous state-of-the-art.
We provide numeric results and additional results when classifying individual object bounding boxes~\cite{borji2020reobjectnet} in the Supplementary Material section~\ref{sec:objectnet-detailed-results}.

\subsection{Object Detection}

Finally, we evaluate \name{} on object detection.
We use the COCO-2017 dataset~\cite{lin2014microsoft} and train a top-performing object detector, RetinaNet~\cite{lin2017focal}, using pre-trained \name{} models as backbones.
Due to memory constraints, we use the ResNet-101x3 architecture for all of our BiT models.
We fine-tune the detection models on the COCO-2017 train split and report results on the validation split using the standard metric~\cite{lin2014microsoft} in Table~\ref{table:detection}.
Here, we do not use \hyper{}, but stick to the standard RetinaNet training protocol, see the Supplementary Material
\begin{wraptable}{r}{6.5cm}\vspace{-13pt}
\begin{tabular}{|c|c|c|}
  \hline
  \bf Model & \bf Upstream data & \bf AP \\
  \hline
  RetinaNet~\cite{lin2017focal} & ILSVRC-2012 & 40.8 \\
  \hline
  RetinaNet (BiT-S) & ILSVRC-2012 & 41.7 \\
  RetinaNet (BiT-M) & ImageNet-21k & 43.2 \\
  RetinaNet (BiT-L) &  JFT-300M  & \bf43.8 \\
  \hline
\end{tabular}
\caption{Object detection performance on COCO-2017~\cite{lin2014microsoft} validation data of RetinaNet models with pre-trained BiT backbones and the literature baseline.}\label{table:detection}\vspace{-10pt}
\end{wraptable}
section~\ref{sec:detection-sup} for details.
Table~\ref{table:detection} demonstrates that BiT models outperform standard ImageNet pre-trained models.
We can see clear benefits of pre-training on large data beyond ILSVRC-2012: pre-training on ImageNet-21k results in a 1.5 point improvement in Average Precision (AP), while pre-training on JFT-300M further improves performance by 0.6 points.

\section{Analysis}
\label{sec:detailed_analysis}

We analyse various components of \name{}: we demonstrate the importance of model capacity, discuss practical optimization caveats and choice of normalization layer.

\subsection{Scaling Models and Datasets}
\label{sec:architecture}

\begin{figure}[b]\vspace{-10pt}
\centering
\includegraphics[width=0.335\textwidth]{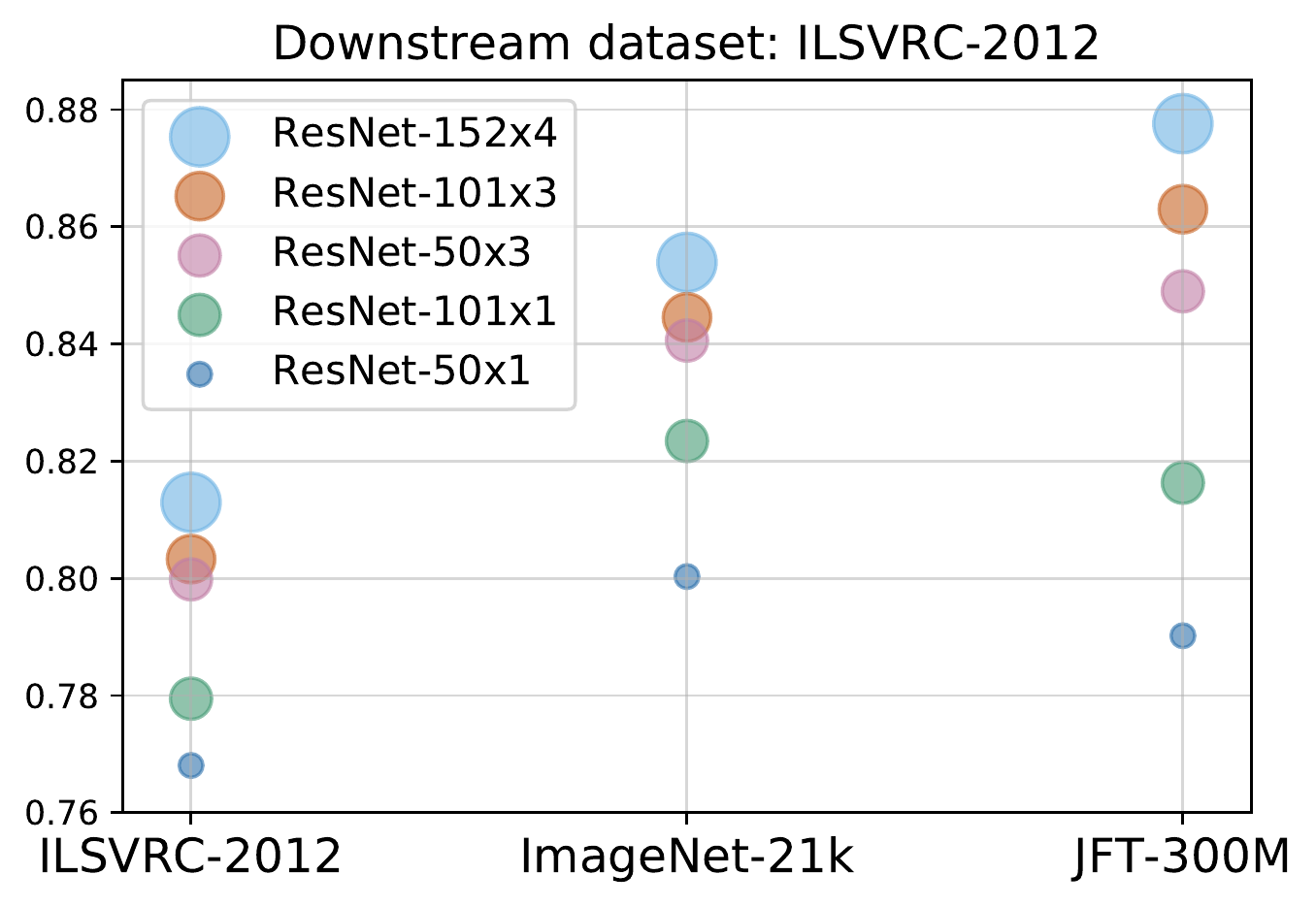}
\!\!\!\!\!
\includegraphics[width=0.335\textwidth]{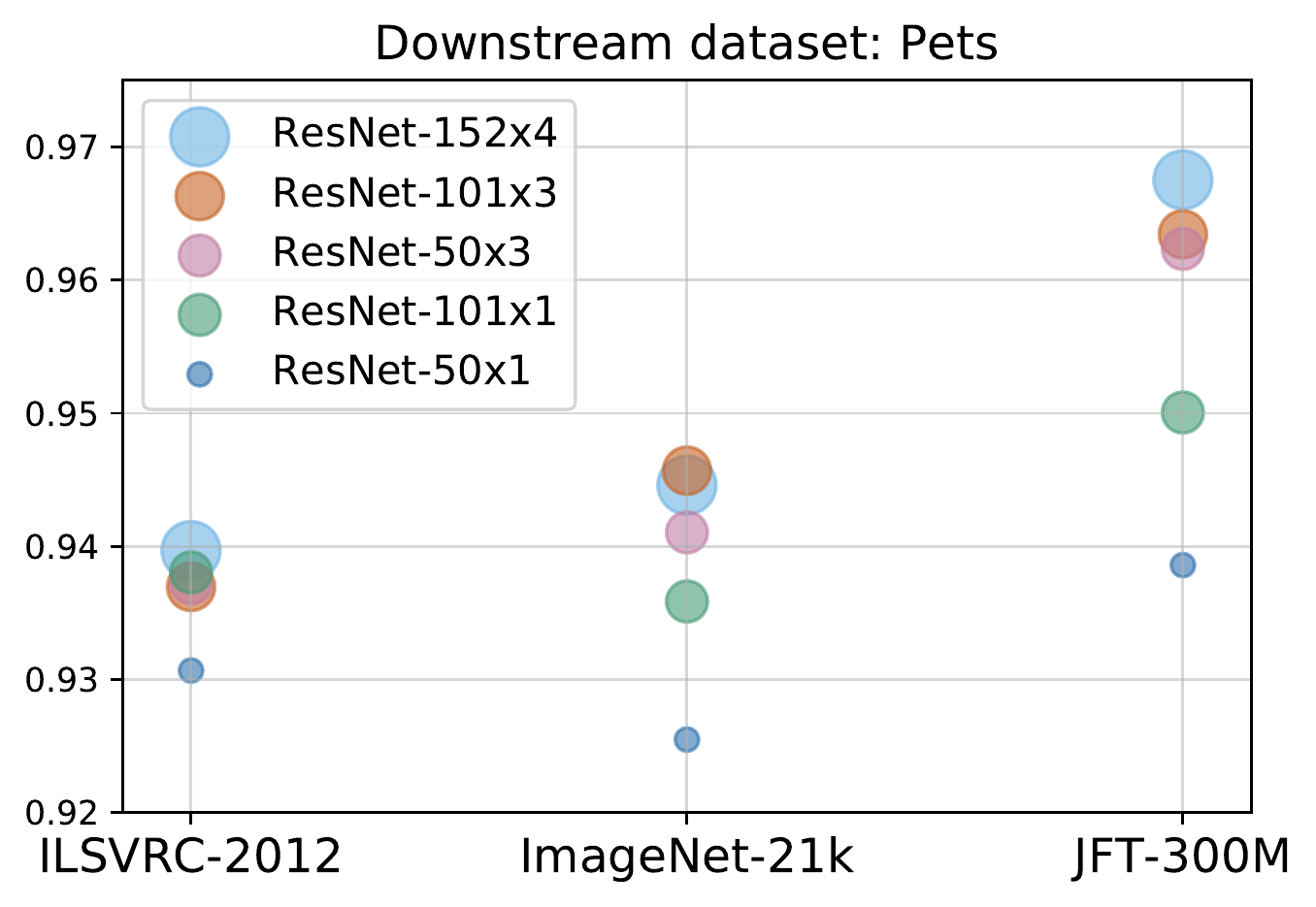}
\!\!\!\!\!
\includegraphics[width=0.335\textwidth]{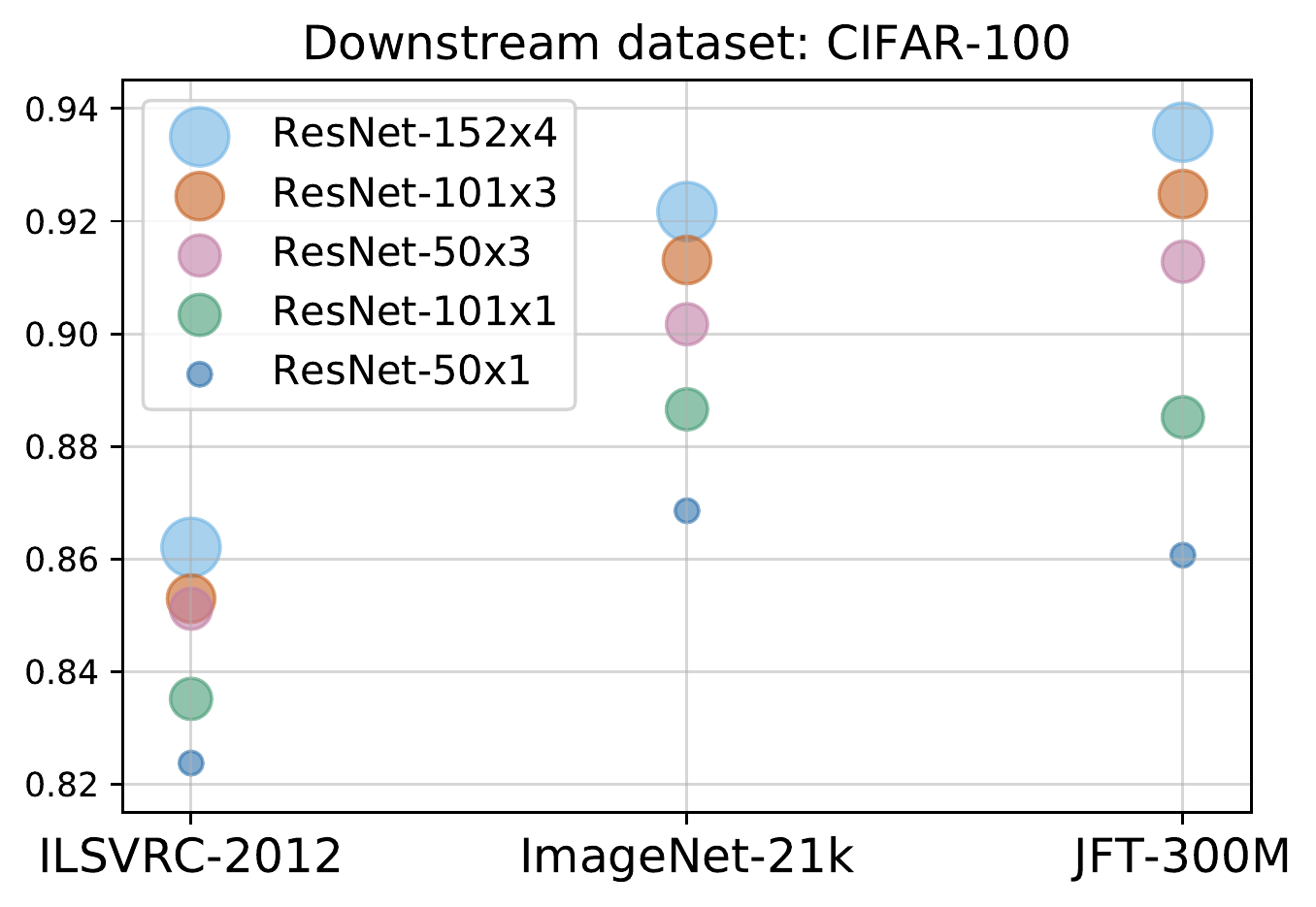}
\caption{Effect of upstream data (shown on the x-axis) and model size on downstream performance. Note that exclusively using more data or larger models may hurt performance; instead, both need to be increased in tandem.}
\label{fig:size_vs_acc}
\end{figure}

The general consensus is that larger neural networks result in better performance.
We investigate the interplay between model capacity and upstream dataset size on downstream performance.
We evaluate the \name{} models of different sizes (ResNet-50x1, ResNet-50x3, ResNet-101x1, ResNet-101x3, and ResNet-152x4) trained on \imagenet{}, ImageNet-21k, and JFT-300M on various downstream benchmarks.
These results are summarized in Figure~\ref{fig:size_vs_acc}.

When pre-training on \imagenet{}, the benefit from larger models diminishes.
However, the benefits of larger models are more pronounced on the larger two datasets.
A similar effect is observed when training on Instagram hashtags~\cite{mahajan2018exploring} and in language modelling~\cite{kaplan2020scaling}.

Not only is there limited benefit of training a large model size on a small dataset, but there is also limited (or even negative) benefit from training a small model on a larger dataset. 
Perhaps surprisingly, the ResNet-50x1 model trained on the JFT-300M dataset can even performs worse than the same architecture trained on the smaller ImageNet-21k.
Thus, if one uses only a ResNet50x1, one may conclude that scaling up the dataset does not bring any additional benefits.
However, with larger architectures, models pre-trained on JFT-300M significantly outperform those pre-trained on \imagenet{} or ImageNet-21k.

\begin{figure}[t]
\centering
\includegraphics[width=0.99\textwidth]{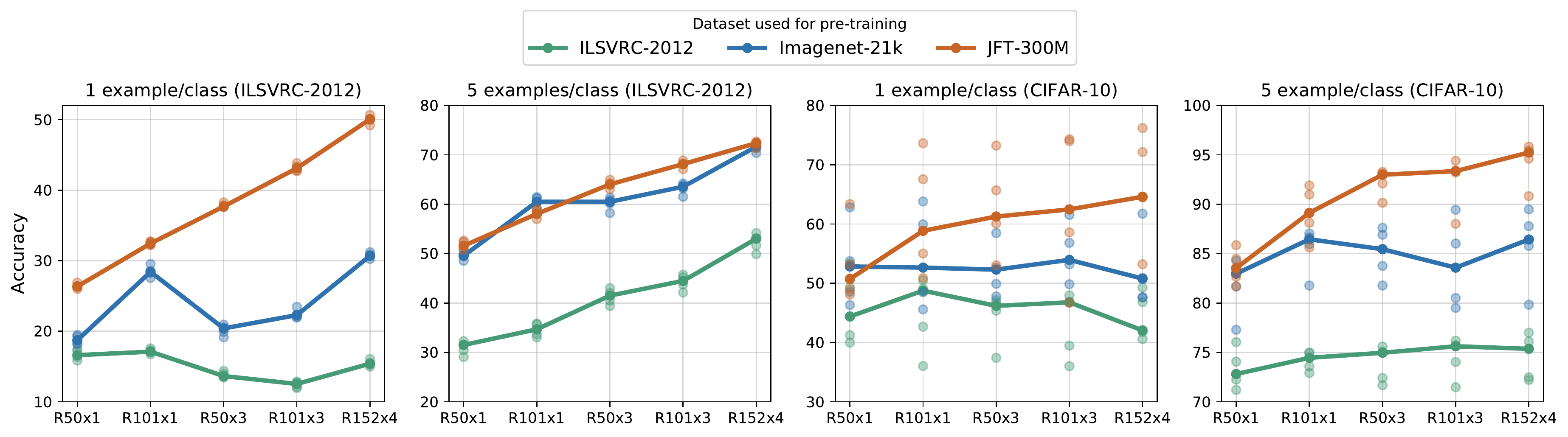}
\caption{Performance of BiT models in the low-data regime. 
The x-axis corresponds to the architecture, where R is short for ResNet.
We pre-train on the three upstream datasets and evaluate on two downstream datasets: ILSVRC-2012 (left) and CIFAR-10 (right) with 1 or 5 examples per class.
For each scenario, we train 5 models on random data subsets, represented by the lighter dots.
The line connects the medians of these five runs. 
}
\label{fig:few_shot_arch}
\end{figure}

Figure~\ref{fig:low_data_plot} shows that \name{}-L attains strong results even on tiny downstream datasets.
Figure~\ref{fig:few_shot_arch} ablates few-shot performance across different pre-training datasets and architectures.
In the extreme case of one example per class, larger architectures outperform smaller ones when pre-trained on large upstream data.
Interestingly, on \imagenet{} with few shots, \name{}-L trained on JFT-300M outperforms the models trained on the entire \imagenet{} dataset itself. 
Note that for comparability, the classifier head is re-trained from scratch during fine-tuning, even when transferring \imagenet{} full to \imagenet{} few shot.

\subsection{Optimization on Large Datasets}
\label{sec:schedule}

\begin{figure}[t]
\centering
\includegraphics[width=\textwidth]{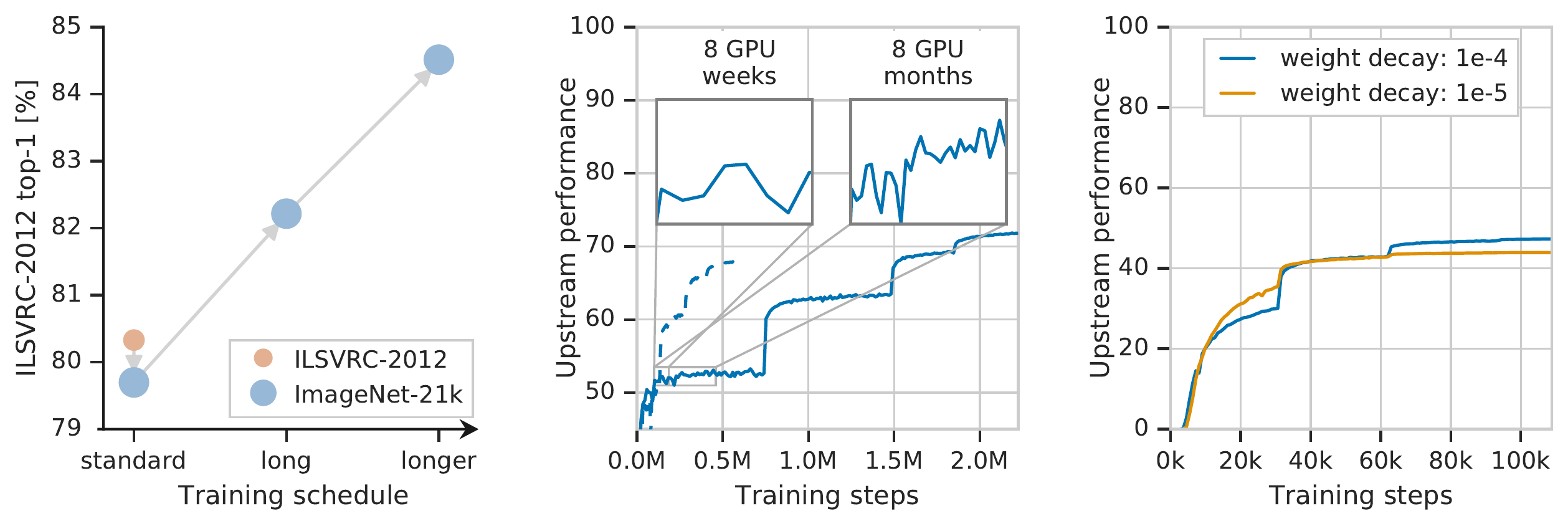}
\caption{
\textbf{Left:} Applying the ``standard'' computational budget of \imagenet{} to the larger ImageNet-21k seems detrimental.
Only when we train longer ($3$x and $10$x) do we see the benefits of training on the larger dataset.
\textbf{Middle:} The learning progress of a ResNet-101x3 on JFT-300M seems to be flat even after 8 GPU-weeks, but after 8 GPU-months progress is clear. If one decays the learning rate too early (dashed curve), final performance is significantly worse.
\textbf{Right}: Faster initial convergence with lower weight decay may trick the practitioner into selecting a sub-optimal value. Higher weight decay converges more slowly, but results in a better final model.
}
\label{fig:schedule}
\end{figure}

For standard computer vision datasets such as \imagenet{}, there are well-known training procedures that are robust and lead to good performance.
Progress in high-performance computing has made it feasible to learn from much larger datasets, such as ImageNet-21k, which has 14.2M images compared to \imagenet{}'s 1.28M. 
However, there are no established procedures for training from such large datasets. 
In this section we provide some guidelines.

Sufficient computational budget is crucial for training performant models on large datasets. The standard \imagenet{} training schedule processes roughly 100 million images (1.28M images $\times$ 90 epochs).
However, if the same computational budget is applied to ImageNet-21k, the resulting model performs worse on \imagenet{}, see Figure~\ref{fig:schedule}, left.
Nevertheless, as shown in the same figure, by increasing the computational budget, we not only recover \imagenet{} performance, but significantly outperforms it.
On JFT-300M the validation error may not improve over a long time ---Figure~\ref{fig:schedule} middle plot, ``8 GPU weeks'' zoom-in--- although the model is still improving as evidenced by the longer time window.

Another important aspect of pre-training with large datasets is the weight decay.
Lower weight decay can result in an apparent acceleration of convergence, Figure~\ref{fig:schedule} rightmost plot.
However, this setting eventually results in an under-performing final model.
This counter-intuitive behavior stems from the interaction of weight decay and normalization layers~\cite{laarhoven17b,li2019exponential}.
Low weight decay results in growing weight norms, which in turn results in a diminishing effective learning rate. Initially this effect creates an impression of faster convergence, but it eventually prevents further progress.
A sufficiently large weight decay is required to avoid this effect, and throughout we use $10^{-4}$.

Finally, we note that in all of our experiments we use stochastic gradient descent with momentum without any modifications.
In our preliminary experiments we did not observe benefits from more involved adaptive gradient methods. 

\subsection{Large Batches, Group Normalization, Weight Standardization}
\label{sec:gn}

Currently, training on large datasets is only feasible using many hardware accelerators.
Data parallelism is the most popular distribution strategy, and this naturally entails large batch sizes.
Many known algorithms for training with large batch sizes use Batch Normalization (BN)~\cite{ioffe2015batch} as a component~\cite{Goyal2017AccurateLM} or even highlight it as the key instrument required for large batch training~\cite{de2019bn}. 

Our larger models have a high memory requirement for any single accelerator chip, which necessitates small per-device batch sizes.
However, BN performs worse when the number of images on each accelerator is too low~\cite{ioffe2017renorm}.
An alternative strategy is to accumulate BN statistics across all of the accelerators.
However, this has two major drawbacks.
First, computing BN statistics across large batches has been shown to harm generalization~\cite{de2019bn}.
Second, using global BN requires many aggregations across accelerators which incurs significant latency.

We investigated Group Normalization (GN)~\cite{wu2018group} and Weight Standardization (WS)~\cite{qiao2019weight} as alternatives to BN.
We tested large batch training using 128 accelerator chips and a batch size of 4096.
We found that GN alone does not scale to large batches; we observe a performance drop of $5.4\%$ on \imagenet{} top-1 accuracy compared to using BN in a ResNet-50x1.
The addition of WS enables GN to scale to such large batches, even outperforming BN, see Table~\ref{tbl:bngn_ilsvrc}.

We are not only interested in upstream performance, but also how models trained with GN and WS transfer.
We thus transferred models with different combinations of BN, GN, and WS pre-trained on \imagenet{} to the 19 tasks defined by VTAB.
The results in Table~\ref{tbl:bngn_vtab} indicate that the GN/WS combination transfers better than BN, so we use GN/WS in all \name{} models.

\begin{table}[t]
  \setlength{\tabcolsep}{0pt}
  \setlength{\extrarowheight}{5pt}
  \renewcommand{\arraystretch}{0.75}
  \centering
  
  \begin{minipage}[t][][b]{0.48\linewidth}
    \centering
    \caption{Top-1 accuracy of ResNet-50 trained from scratch on \imagenet{} with a batch-size of 4096.}\label{tbl:bngn_ilsvrc}
    \begin{tabularx}{\linewidth}{p{2.0cm}p{0.2cm}Cp{0.2cm}C}
      \toprule[1pt]
       && Plain Conv && Weight Std. \\
      \midrule
      Batch Norm. && 75.6 && 75.8 \\
      Group Norm. && 70.2 && \bf 76.0 \\
      \bottomrule
    \end{tabularx}
  \end{minipage}\hfill
  \begin{minipage}[t][][b]{0.48\linewidth}
    \centering
    \caption{Transfer performance of the corresponding models from Table~\ref{tbl:bngn_ilsvrc} fine-tuned to the 19 VTAB-1k tasks.}\label{tbl:bngn_vtab}
    \begin{tabularx}{\linewidth}{p{2.0cm}p{0.2cm}Cp{0.2cm}C}
      \toprule[1pt]
       && Plain Conv && Weight Std. \\
      \midrule
      Batch Norm. && 67.72 && 66.78 \\
      Group Norm. && 68.77 && \bf 70.39 \\
      \bottomrule
    \end{tabularx} \label{table:gnws}
  \end{minipage}\vspace{-10pt}
\end{table}

\section{Related Work}
\label{sec:related}

\subsubsection{Large-scale Weakly Supervised Learning of Representations}
A number of prior works  use large supervised datasets for pre-training visual representations~\cite{joulin2016learning,sun2017revisiting,li2017learning,mahajan2018exploring}.
In \cite{joulin2016learning,li2017learning} the authors use a dataset containing 100M Flickr images~\cite{thomee2015yfcc100m}.
This dataset appears to transfer less well than JFT-300M.
While studying the effect of dataset size, \cite{sun2017revisiting} show good transfer performance when training on JFT-300M, despite reporting a large degree of noise (20\% precision errors) in the labels.
An even larger, noisily labelled dataset of 3.5B Instagram images is used in \cite{mahajan2018exploring}.
This increase in dataset size and an improved model architecture~\cite{xie2017aggregated} lead to better results when transferring to \imagenet{}.
We show that we can attain even better performance with ResNet using JFT-300M with appropriate adjustments presented in Section~\ref{sec:methods}.
The aforementioned papers focus on transfer to ImageNet classification, and COCO or VOC detection and segmentation.
We show that transfer is also highly effective in the low data regime, and works well on the broader set of 19 tasks in VTAB~\cite{zhai2019visual}.

\vspace{-2mm}

\subsubsection{Specialized Representations}
Rather than pre-train generic representations, recent works have shown strong performance by training task-specific representations~\cite{yalniz2019billion,dat,noisystudent}.
These papers condition on a particular task when training on a large support dataset.
\cite{yalniz2019billion,noisystudent} train student networks on a large unlabelled support dataset using the predictions of a teacher network trained on the target task.
\cite{dat} compute importance weights on the a labelled support dataset by conditioning on the target dataset.
They then train the representations on the re-weighted source data.
Even though these approaches may lead to superior results, they require knowing the downstream dataset in advance and substantial computational resources for each downstream dataset.

\begin{figure}[t]
\centering
\begin{subfigure}[b]{0.66\textwidth}
  \centering
  \includegraphics[width=\textwidth]{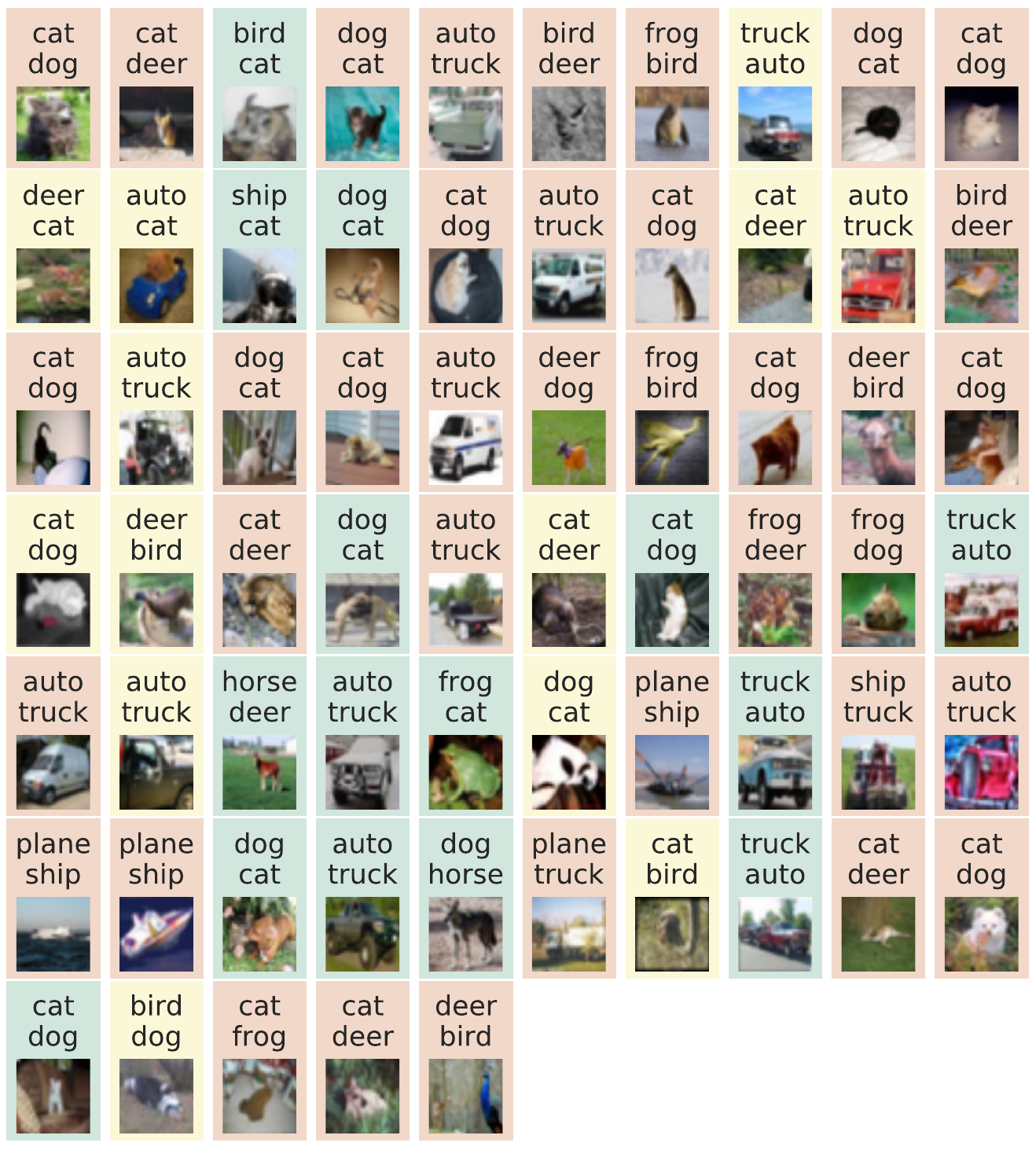}
\end{subfigure}
\hfill
\begin{subfigure}[b]{0.29\textwidth}
 \centering
 \includegraphics[width=\textwidth]{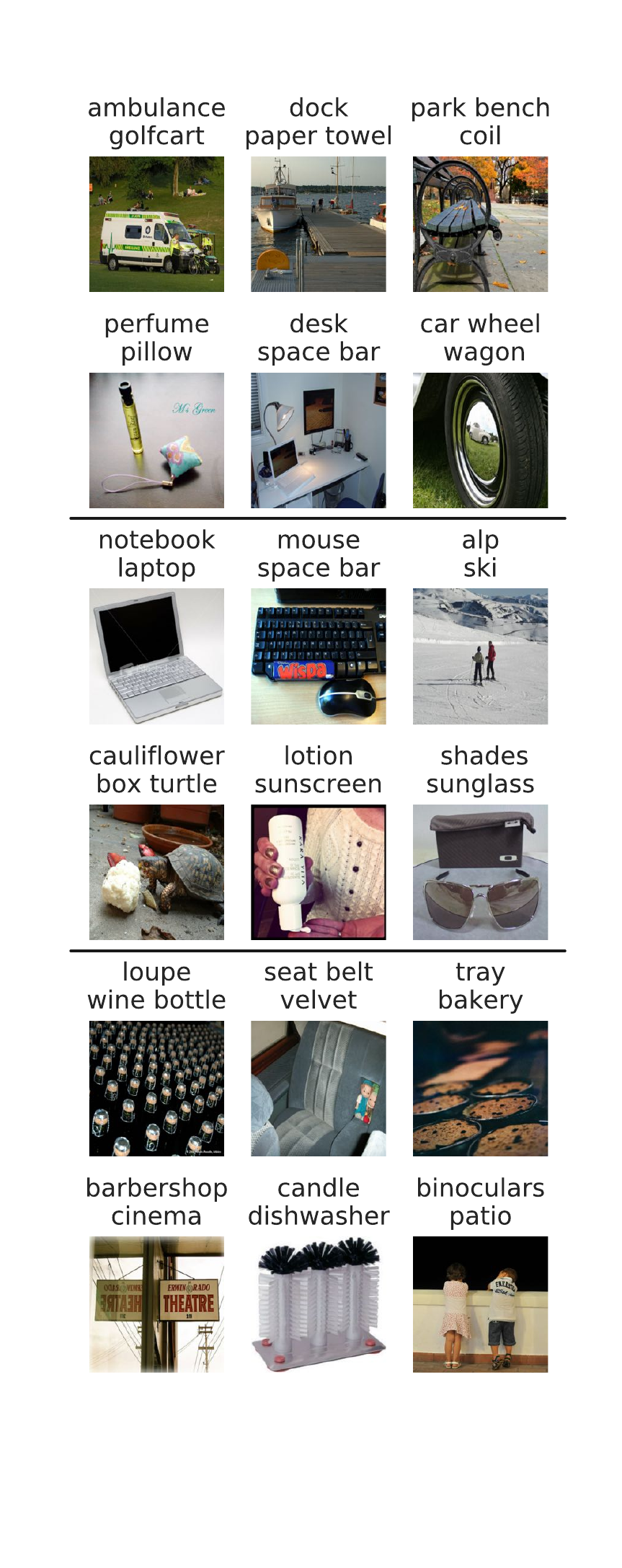}
\end{subfigure}
\caption{
Cases where \name{}-L's predictions (top word) do not match the ground-truth labels (bottom word), and hence are counted as top-1 errors. 
\textbf{Left:} 
\emph{All} mistakes on CIFAR-10, colored by whether five human raters agreed with \name{}-L's prediction (green), with the ground-truth label (red) or were unsure or disagreed with both (yellow).
\textbf{Right:} 
Selected representative mistakes of \name{}-L on \imagenet{}.
Top group: The model's prediction is more representative of the primary object than the label.
Middle group: According to top-1 accuracy the model is incorrect, but according to top-5 it is correct.
Bottom group: The model's top-10 predictions are incorrect.}\label{fig:mistakes}\vspace{-10pt}
\end{figure}

\vspace{-2mm}

\subsubsection{Unsupervised and Semi-Supervised Representation learning}
Self-su\-per\-vi\-sed methods have shown the ability to leverage unsupervised datasets to transfer to labelled tasks.
For example, \cite{he2019momentum} show that unsupervised representations trained on 1B unlabelled Instagram images transfer comparably or better than supervised \imagenet{} features.
Semi-supervised learning exploits unsupervised data drawn from the same domain as the labelled data.
\cite{berthelot2019remixmatch,sohn2020fixmatch} used semi-supervised learning to attain strong performance on CIFAR-10 and SVHN using only 40 or 250 labels.
Recent works combine self-supervised and semi-supervised learning to attain good performance with fewer labels on ImageNet~\cite{zhai2019s4l,henaff2019data}.
\cite{zhai2019visual}~study many representation learning algorithms (unsupervised, semi-su\-per\-vi\-sed, and supervised) and evaluate their representation's ability to generalize to novel tasks, concluding that a combination of supervised and self-supervised signals works best. 
However, all models were trained on \imagenet{}.
We show that supervised pre-training on larger datasets continues to be an effective strategy.

\vspace{-2mm}

\subsubsection{Few-shot Learning}
Many strategies have been proposed to attain good performance when faced with novel classes and only a few examples per class.
Meta-learning or metric-learning techniques have been proposed to learn with few or no labels~\cite{vinyals2016matching,snell2017prototypical,sung2018learning}.
However, recent work has shown that a simple linear classifier on top of pre-trained representations or fine-tuning can attain similar or better performance~\cite{chen2019closerlook,nakamura2019revisiting}.
The upstream pre-training and downstream few-shot learning are usually performed on the same domain, with disjoint class labels. 
In contrast, our goal is to find a generalist representation which works well when transferring to many downstream tasks.

\section{Discussion}

We revisit classical transfer learning, where a large pre-trained generalist model is fine-tuned to downstream tasks of interest.
We provide a simple recipe which exploits large scale pre-training to yield good performance on all of these tasks.
\name{} uses a clean training and fine-tuning setup, with a small number of carefully selected components, to balance complexity and performance.

In Figure~\ref{fig:mistakes} and the Supplementary Material section~\ref{sec:all-mistakes}, we take a closer look at the remaining mistakes that \name{}-L makes.
In many cases, we see that these label/prediction mismatches are not true `mistakes': the model's classification is valid, but it does not match the label.
For example, the model may identify another prominent object when there are multiple objects in the image, or may provide an valid classification when the main entity has multiple attributes.
There are also cases of label noise, where the model's prediction is a better fit than the ground-truth label.
In a quantitative study, we found that around half of the model's mistakes on CIFAR-10 are due to ambiguity or label noise (see Figure~\ref{fig:mistakes}, left), and in only 19.21\% of the \imagenet{} mistakes do human raters clearly agree with the label over the prediction.
Overall, by inspecting these mistakes, we observe that performance on the standard vision benchmarks seems to approach a saturation point. 

We therefore explore the effectiveness of transfer to two classes of more challenging tasks:
classical image recognition tasks, but with very few labelled examples to adapt to the new domain, and VTAB, which contains more diverse tasks, such as spatial localization, tasks from simulated environments, and medical and satellite imaging tasks.
These benchmarks are much further from saturation; while \name{}-L performs well on them, there is still substantial room for further progress.

\section{Acknowledgements}

We thank the whole Google Brain team in Z\"urich and its collaborators for many fruitful discussions and engineering support.
In particular, we thank Andrei Giurgiu for finding a bug in our data input pipeline, Marcin Michalski for the naming idea and general helpful advice, and Damien Vincent and Daniel Keysers for detailed feedback on the initial draft of this paper.

%% file: appendix.tex
\appendix

\section{Tuning hyperparameters for transfer}\label{sec:tuningtransfer}

\begin{figure}[t]
\begin{center}
   \includegraphics[width=1.0\textwidth]{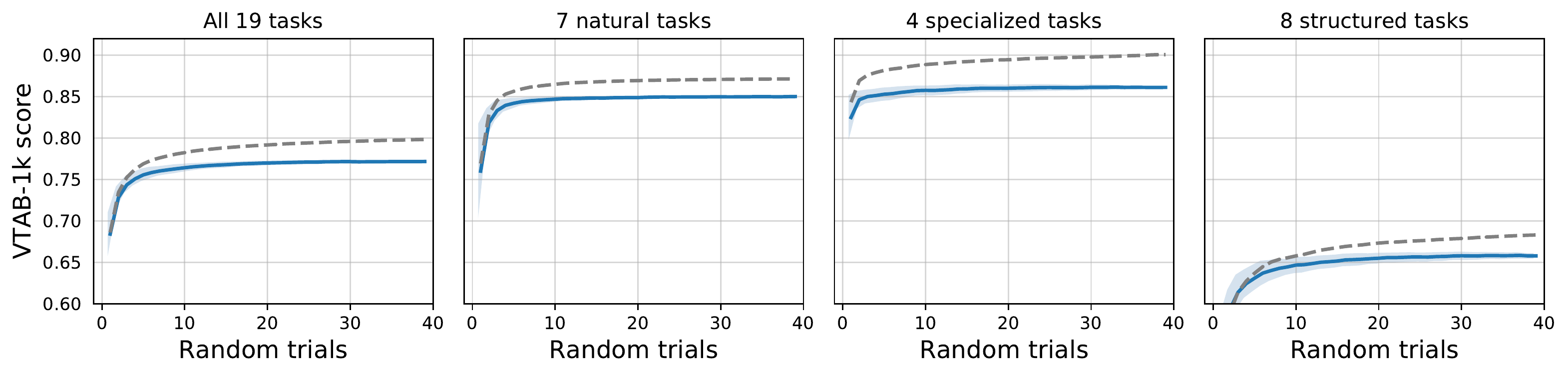}
\end{center}
\caption{Blue curves display VTAB-1k score (mean accuracy across tasks) depending on the total number of random hyperparameters tested. Reported VTAB-1k scores are averaged over 100 random hyperparameter orderings, the shaded blue area indicates the standard error. Dashed gray line displays the performance on the small hold-out validation split with 200 examples.}
\label{fig:budget}
\end{figure}

Throughout the paper we evaluate \name{} using \hyper{}.
Here, we investigate whether \name{}-L would benefit from additional computational budget for selecting fine-tuning hyperparameters.

For this investigation we use VTAB-1k as it contains a diverse set of 19 tasks.
For each task we fine-tune \name{}-L 40 times using 800 training images.
Each trial uses randomly sampled hyperparameters as described below.
We select the best model for each dataset using the validation set with 200 images.
The results are shown in Figure~\ref{fig:budget}.
Overall, we observe that VTAB-1k score saturates roughly after 20 trials and that further tuning results in overfitting on the validation split.
This indicates that practitioners do not need to do very heavy tuning in order to find optimal parameters for their task.

After re-training \name{}-L model with selected hyper-parameters using all union of training and validation splits (1000 images) we obtain the VTAB-1k score of 78.72\%, an absolute improvement of 2.43\% over 76.29\% score obtained with BiT-HyperRule.

Our random search includes following hyperparameters with the following ranges and sampling strategies: 
\begin{itemize}
    \item Initial learning rate is sampled log-uniformly from the range $[10^{-1}, 10^{-4}]$.
    \item Total number of updates is sampled from the set $\{500, 1000, 2000, 4000, 8000, \\ 16000\}$.
    \item Dropout rate for the penultimate layer is uniformly sampled from the range $[0.0, 0.7]$.
    \item Weight decay to the initial weight values is sampled  log-uniformly from the range $[10^{-1}, 10^{-6}]$  .
    \item MixUp $\alpha$ parameter is sampled from the set $\{\text{None}, 0.05, 0.1, 0.2, 0.4\}$.
    \item Input image resolution is sampled from the set $\{64, 128, 192, 256, 320, 384\}$. 
\end{itemize}

\clearpage

\section{Full ObjectNet results}
\label{sec:objectnet-detailed-results}

\begin{figure}
    \begin{subfigure}[t]{0.9\textwidth}
      \includegraphics[width=\textwidth]{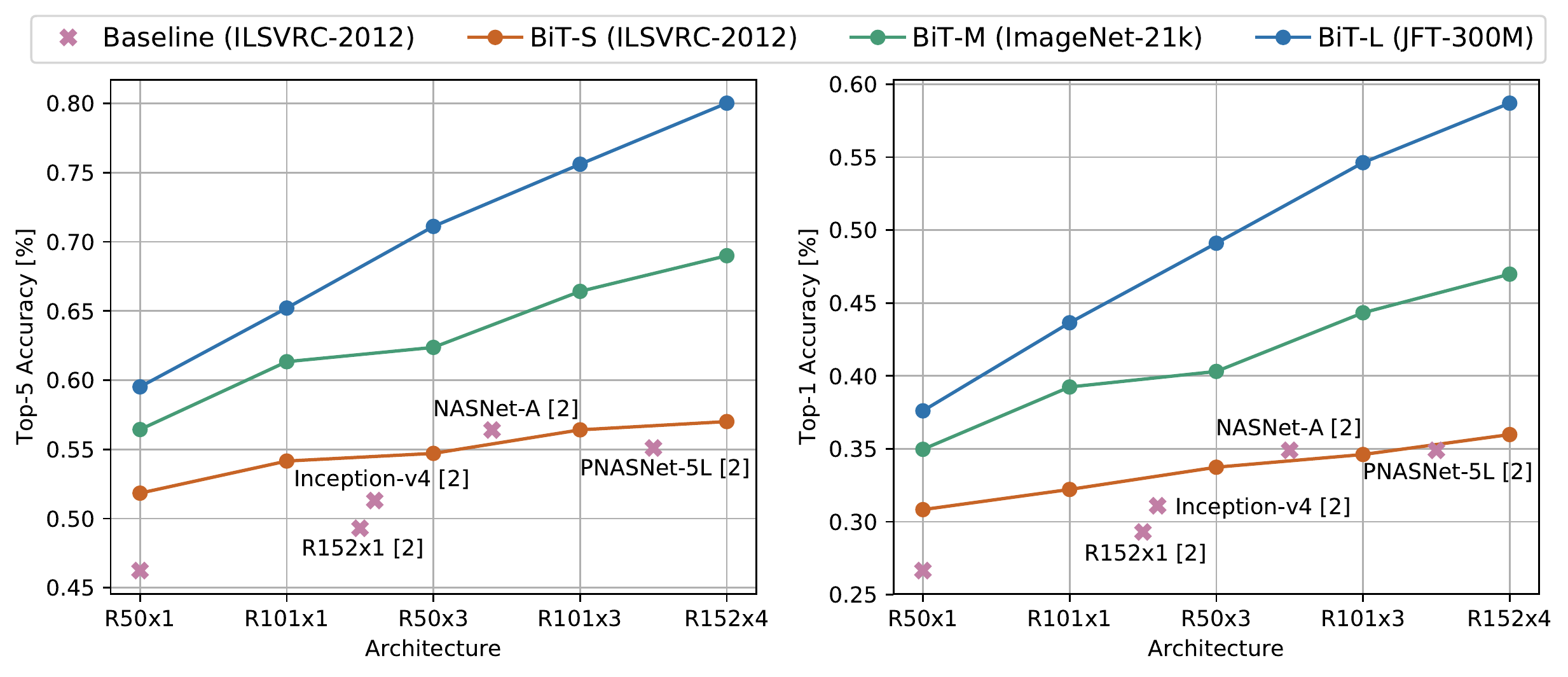}
      \caption{Results on the original ObjectNet test set with resize and central crop.}
    \end{subfigure}
    \hfill
    \begin{subfigure}[t]{0.9\textwidth}
      \includegraphics[width=\textwidth]{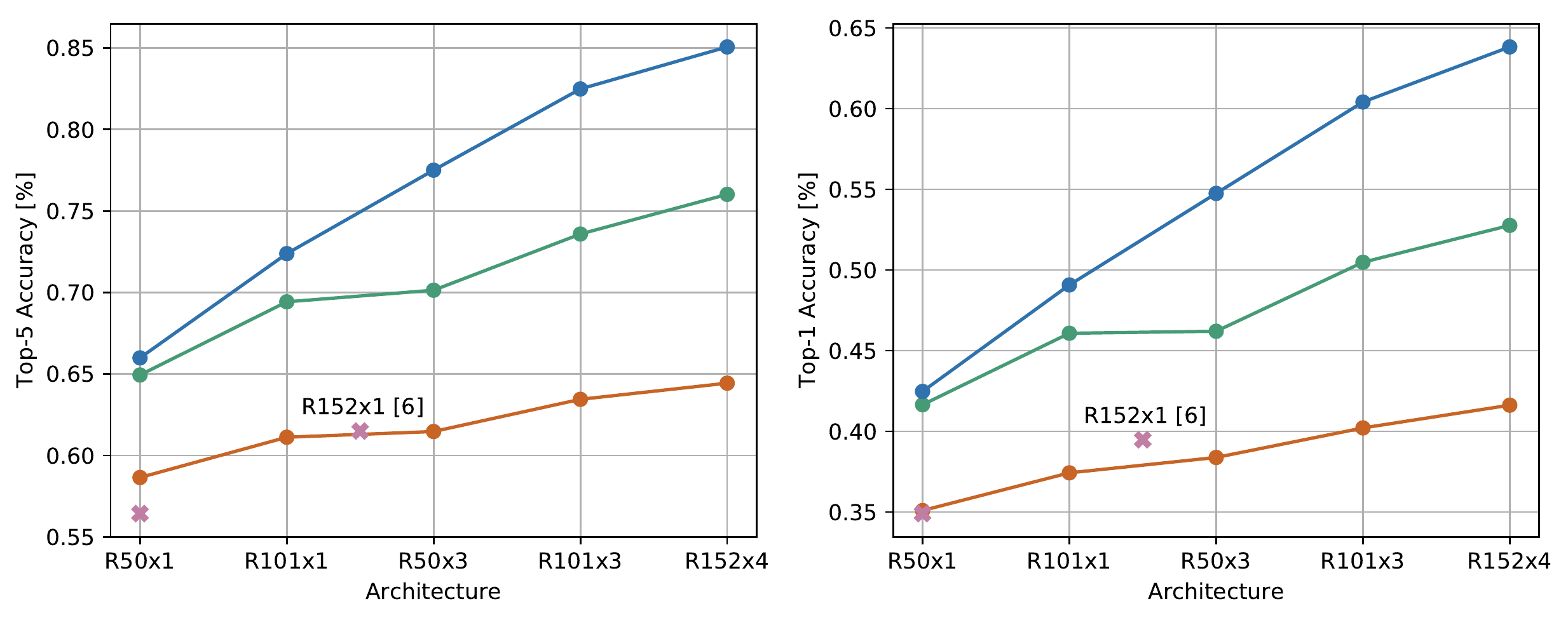}
      \caption{Results on the cropped ObjectNet, where individual objects are cropped for evaluation. The bounding boxes are provided by \cite{borji2020reobjectnet}.}
    \end{subfigure}
    \caption{Results on ObjectNet; left: top-5 accuracy, right: top-1 accuracy.}
\label{fig:objectnet_detailed}
\end{figure}

Figure~\ref{fig:objectnet_detailed} shows more results on the ObjectNet test set, with top-5 accuracy reported on the left and top-1 accuracy on the right.
In Figure~\ref{fig:objectnet_detailed}~(a), we first resize the shorter side of the image to 512 pixels and then take a $480{\times}480$ pixel sized central crop, similar to \hyper{}.

ObjectNet is a dataset collected in the real world, where multiple objects are present most of the time.
A recent analysis shows that cropping out a single object from the cluttered scene could significantly improve performance \cite{borji2020reobjectnet}. 
In Figure~\ref{fig:objectnet_detailed}~(b), we follow the same setup and report our models' performance on the cropped ObjectNet with a single object in each crop.
We observe a solid improvement in performance in this setting.

Overall, the trend of our improvements is consistent with the results on the original ObjectNet test set.
We provide our full numeric results in Table~\ref{tbl:objectnet}.

\begin{table}[t]
  \setlength{\tabcolsep}{0pt}
  \setlength{\extrarowheight}{5pt}
  \renewcommand{\arraystretch}{0.75}
  \centering
  \caption{Results (\%) on the ObjectNet test set. We report numbers for both the standard setting, as well as for the setting where the ground-truth bounding box is used.}
  \label{tbl:objectnet}
  \begin{tabularx}{\linewidth}{
      p{1.5cm}
      p{0.2cm}Cp{0.2cm}Cp{0.2cm}C
      p{0.3cm}Cp{0.2cm}Cp{0.2cm}C
      p{0.4cm}Cp{0.2cm}Cp{0.2cm}C
      p{0.3cm}Cp{0.2cm}Cp{0.2cm}C}
    \toprule[1pt]
     && \multicolumn{11}{c}{Top-1 accuracy} && \multicolumn{11}{c}{Top-5 accuracy} \\
    \cmidrule[0.5pt]{3-13} \cmidrule[0.5pt]{15-25}
     && \multicolumn{5}{c}{Resize \& Crop} && \multicolumn{5}{c}{Bounding Box} && \multicolumn{5}{c}{Resize \& Crop} && \multicolumn{5}{c}{Bounding Box} \\
    \cmidrule[0.5pt]{3-7} \cmidrule[0.5pt]{9-13} \cmidrule[0.5pt]{15-19} \cmidrule[0.5pt]{21-25}
    BiT- && S && M && L && S && M && L && S && M && L && S && M && L \\
    \midrule
R50x1  &&	30.8	&&	35.0	&&	37.6	&&	35.1	&&	41.6	&&	42.5	&&	51.8	&&	56.4	&&	59.5	&&	58.7	&&	64.9	&&	66.0	\\
R101x1 &&	32.2	&&	39.2	&&	54.6	&&	37.4	&&	46.1	&&	49.1	&&	54.2	&&	61.3	&&	75.6	&&	61.1	&&	69.4	&&	72.4	\\
R50x3  &&	33.7	&&	40.3	&&	49.1	&&	38.4	&&	46.2	&&	54.7	&&	54.7	&&	62.4	&&	71.1	&&	61.5	&&	70.1	&&	77.5	\\
R101x3 &&	34.6	&&	44.3	&&	54.6	&&	40.2	&&	50.5	&&	60.4	&&	56.4	&&	66.4	&&	75.6	&&	63.4	&&	73.6	&&	82.5	\\
R152x4 &&	36.0	&&	47.0	&&	\textbf{58.7}	&&	41.6	&&	52.8	&&	\textbf{63.8}	&&	57.0	&&	69.0	&&	\textbf{80.0}	&&	64.4	&&	76.0	&&	\textbf{85.1}	\\
    \bottomrule
  \end{tabularx}
\end{table}

\clearpage

\section{Duplicates and near-duplicates}\label{sec:dedup}

\begin{table}[t]
  \setlength{\tabcolsep}{0pt}
  \setlength{\extrarowheight}{5pt}
  \renewcommand{\arraystretch}{0.75}
  \centering
  \caption{Performance of \name{}-L on the original (``Full'') and deduplicated (``Dedup'') test data. The ``Dups'' column shows the total  number of near-duplicates found.}
  \label{tbl:dedup}
  \begin{tabularx}{\linewidth}{
      p{3.0cm}
      p{0.2cm}Cp{0.2cm}Cp{0.2cm}S[table-number-alignment = right, table-figures-decimal = 0]
      p{0.2cm}Cp{0.2cm}Cp{0.2cm}S[table-number-alignment = right, table-figures-decimal = 0]
      p{0.2cm}Cp{0.2cm}Cp{0.2cm}S[table-number-alignment = right, table-figures-decimal = 0]}
    \toprule[1pt]
     && \multicolumn{5}{c}{From JFT} && \multicolumn{5}{c}{From ImageNet21k} && \multicolumn{5}{c}{From \imagenet{}} \\
    \cmidrule[0.5pt]{3-7} \cmidrule[0.5pt]{9-13} \cmidrule[0.5pt]{15-19}
     && Full && Dedup && {Dups} && Full && Dedup && {Dups} && Full && Dedup && {Dups} \\
    \midrule

    \imagenet{} && 87.8 && 87.9 && 6470 && 84.5 && 85.3 && 3834 && 80.3 && 81.3 && 879 \\
    CIFAR-10    && 99.4 && 99.3 &&  435 && 98.5 && 98.4 &&  687 && 97.2 && 97.2 &&  82 \\
    CIFAR-100   && 93.6 && 93.4 &&  491 && 91.2 && 90.7 &&  890 && 85.3 && 85.2 && 136 \\
    Pets        && 96.8 && 96.4 &&  600 && 94.6 && 94.5 &&   80 && 93.7 && 93.6 &&  58 \\
    Flowers     && 99.7 && 99.7 &&  412 && 99.5 && 99.5 &&  335 && 91.0 && 91.0 &&   0 \\
    \bottomrule
  \end{tabularx}
\end{table}

In order to make sure that our results are not inflated due to overlap between upstream training and downstream test data, we run extensive de-duplication experiments.
For training our flagship model, \name{}-L, we remove all images from JFT-300M dataset that are duplicates and near-duplicates of test images of all our downstream datasets. In total, we removed less than 50\,k images from the JFT-300M dataset.
Interestingly, we did not observe any drastic difference by doing de-duplication, evidenced by comparing the first column of Table~\ref{tbl:main} (de-duplicated upstream) and the first column of Table~\ref{tbl:dedup} (full upstream).

In another realistic setting, eventual downstream tasks are not known in advance.
To better understand this setting, we also investigate how duplicates affect performance by removing them from the downstream test data after the upstream model has already been trained.
The results of this experiment are shown in Table~\ref{tbl:dedup}: ``Full'' is the accuracy on the original test set that contains near-duplicates, ``Dedup'' is the accuracy on the test set cleaned of near-duplicates, and ``Dups'' is the number of near-duplicates that have been removed from said test set.
We observe that near-duplicates barely affect the results in all of our experiments.
Note that near-duplicates between training and test sets have previously been reported by \cite{sun2017revisiting} for \imagenet{}, and by \cite{barz2019duplicates} for CIFAR.

\begin{figure}[t]
\begin{center}
   \includegraphics[width=1.0\textwidth]{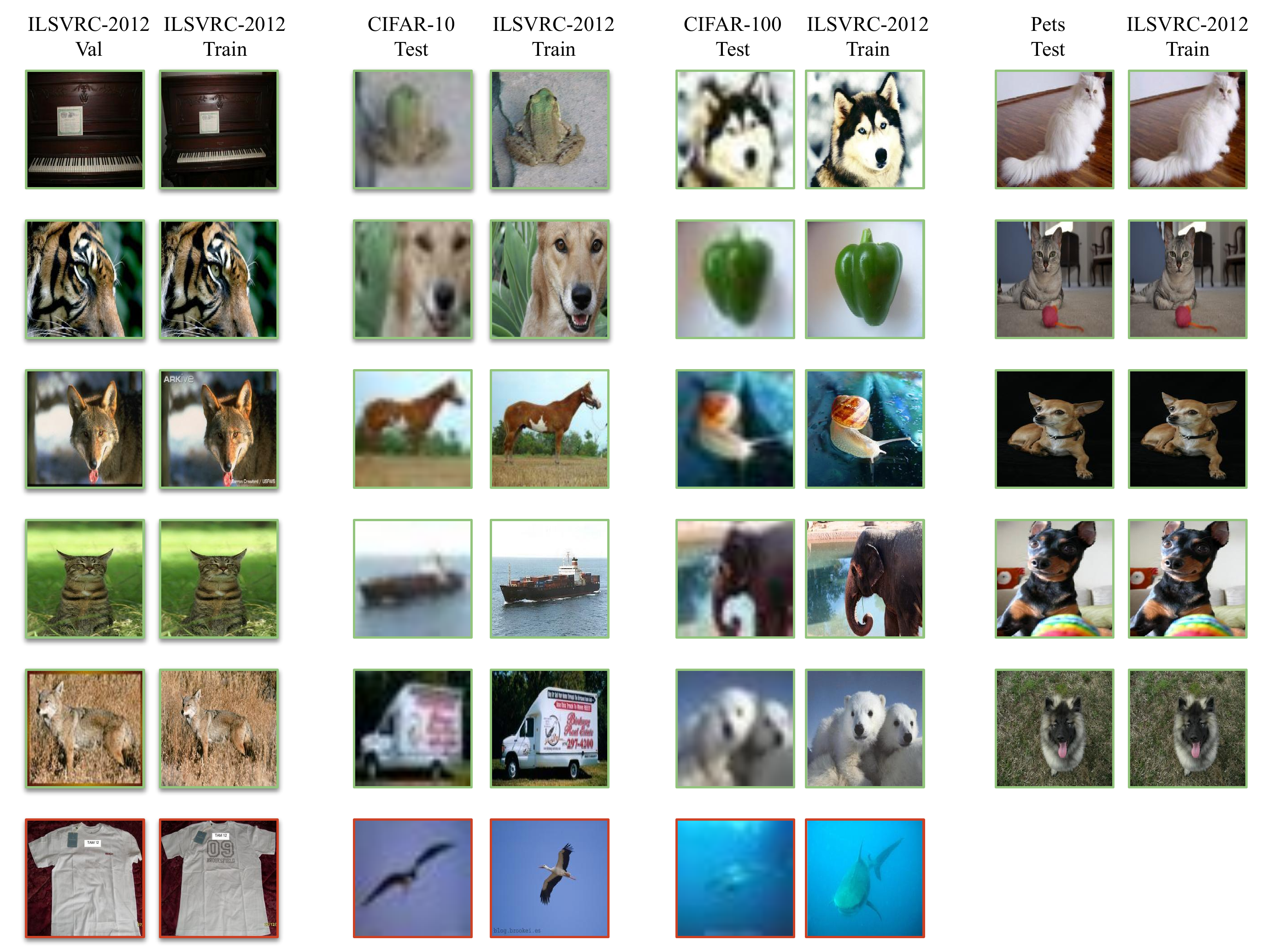}
\end{center}
\caption{Detected duplicates between the \imagenet{} training set and test splits of various downstream datasets. Note that Flowers is not listed because there are no duplicates. Green borders mark true positives and red borders mark (rare) false positives.}
\label{fig:duplicates}
\end{figure}

In Figure~\ref{fig:duplicates}, we present a few duplicates found between the ILSVRC-2012 training set and test splits of four standard downstream datasets.

\clearpage

\section{All of \name{}-L's Mistakes}\label{sec:all-mistakes}

\begin{figure}[ht]
\centering
\includegraphics[width=\textwidth]{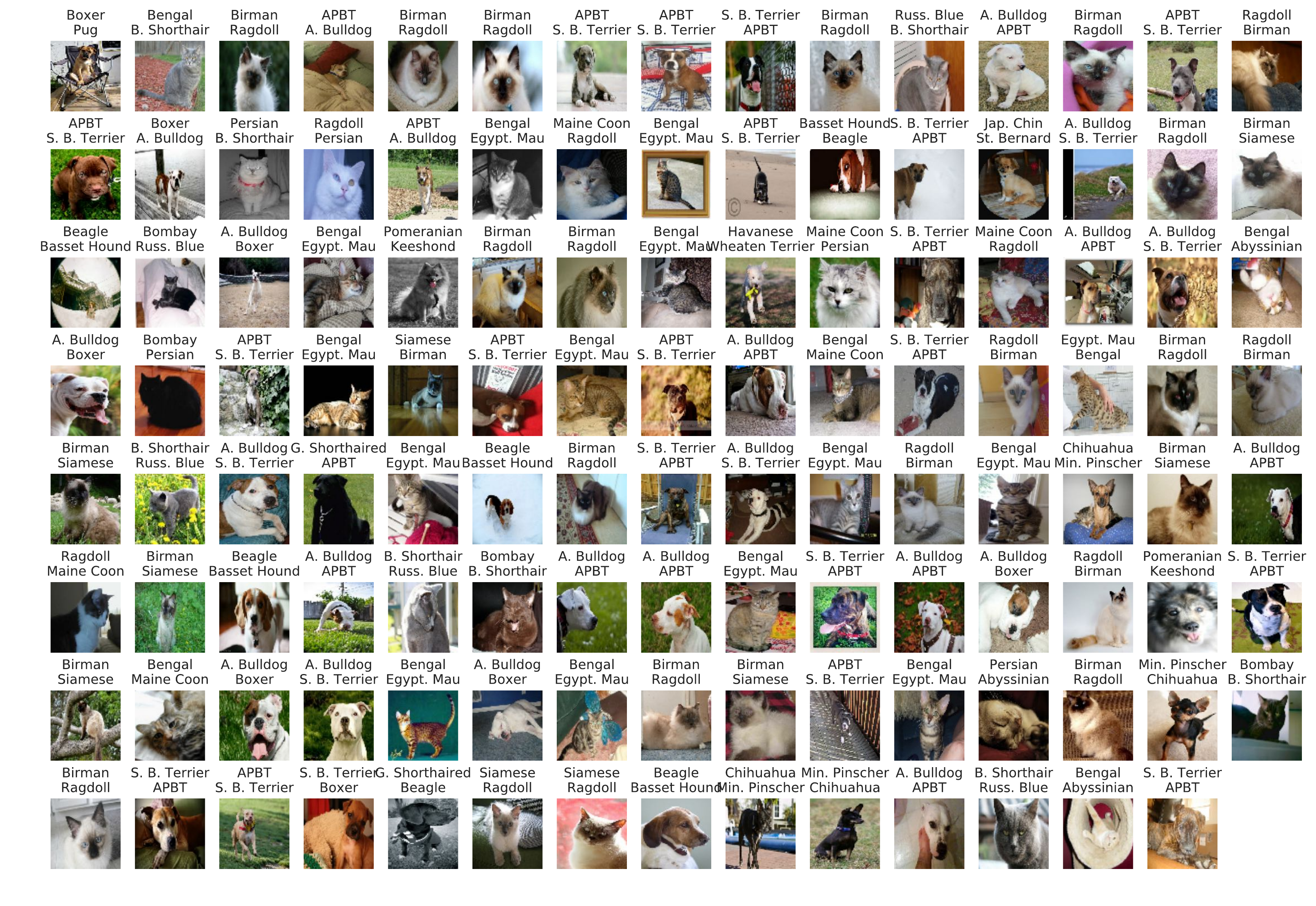}
\caption{All of \name{}-L's mistakes on Oxford-IIIT-Pet.}\label{fig:mistakes-pets}
\end{figure}

\begin{figure}[ht]
\centering
\includegraphics[width=0.9\textwidth]{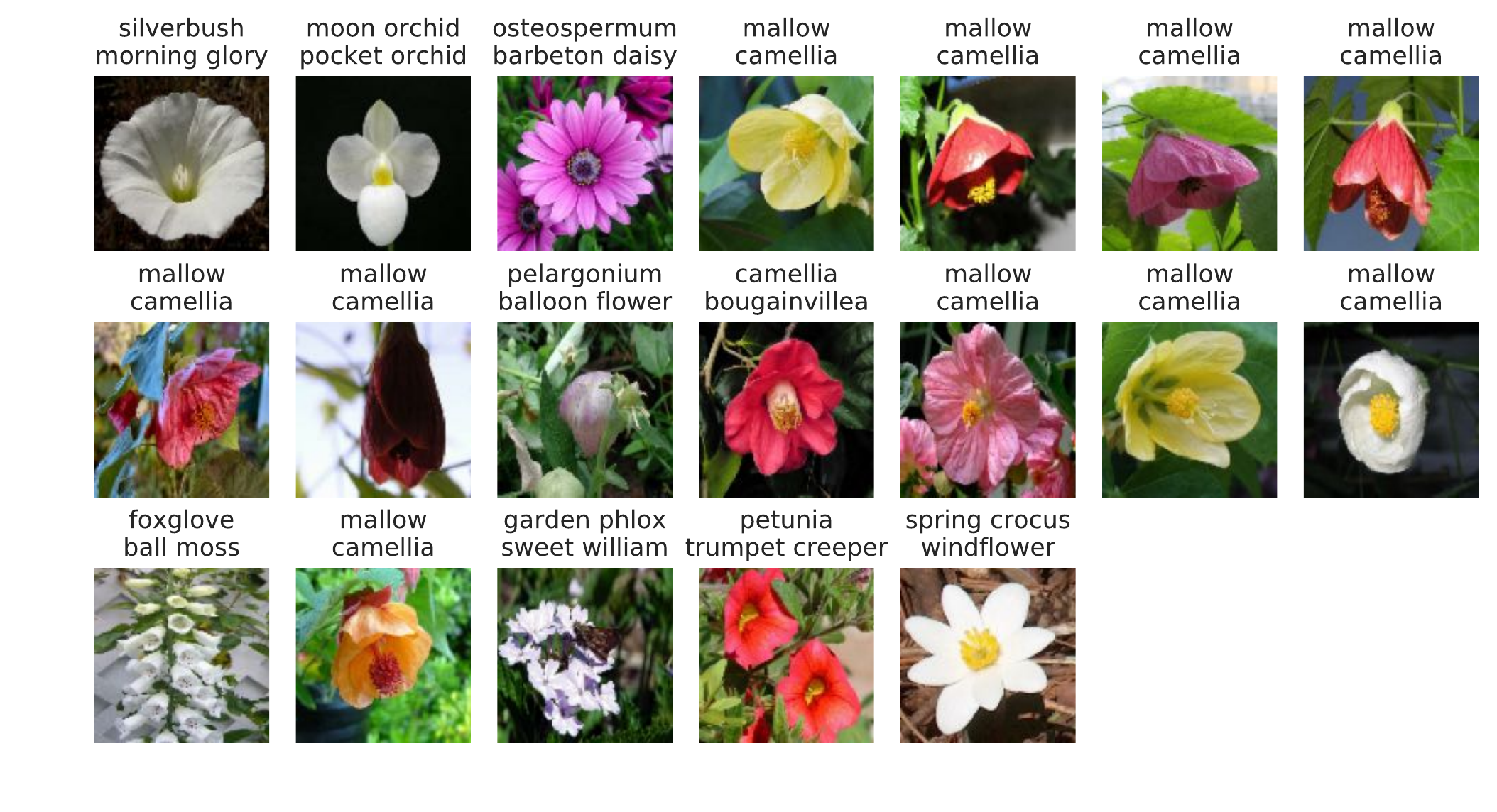}
\caption{All of \name{}-L's mistakes on Oxford-Flowers102.}\label{fig:mistakes-flowers}
\end{figure}

Here we take a closer look at the mistakes made by \name{}-L\footnote{To be precise, the figures are obtained by an earlier version of our \name{}-L model but which reaches almost the same accuracy. We did not re-run the figures and human evaluation with the latest model as they serve for illustration purposes and the models perform essentially the same, modulo a few flips.}.
Figure~\ref{fig:mistakes}, we show \emph{all} mistakes on CIFAR-10, as well as a representative selection of mistakes on ILSVRC-2012.
Figures~\ref{fig:mistakes-pets} and~\ref{fig:mistakes-flowers} again show \emph{all} mistakes on the Pets and Flowers datasets, respectively.
The first word always represents the model's prediction, while the second word represents the ground-truth label.
The larger panels are best viewed on screen, where they can be magnified.

\clearpage

\section{Object detection experiments}\label{sec:detection-sup}

As discussed in the main text, for object detection evaluation we use the RetinaNet model~\cite{lin2017focal}.
Our implementation is based on publicly available code\footnote{\url{https://github.com/tensorflow/tpu/tree/master/models/official/retinanet}} and uses standard hyper-parameters for training all detection models.
We repeat training 5 times and report median performance. 

Specifically, we train all of our models for 30 epochs using a batch size of 256 with stochastic gradient descent, 0.08 initial learning rate, 0.9 momentum and $10^{-4}$ weight decay.
We decrease the initial learning rate by a factor of 10 at epochs number 16 and 22.
We did try training for longer (60 epochs) and did not observe performance improvements.
The input image resolution is $1024 \times 1024$.
During training we use a data augmentation scheme as in~\cite{lin2014microsoft}: random horizontal image flips and scale jittering.
We set the classification loss parameters $\alpha$ to 0.25 and $\gamma$ to 2.0, see~\cite{lin2017focal} for the explanation of these parameters.

\section{Horizontal flipping and cropping for VTAB-1k tasks}\label{sec:flip-and-crop-details}

When fine-tuning \name{} models, we apply random horizontal flipping and cropping as image augmentations. 
However, these operations are not reasonable for certain VTAB tasks, where the semantic label (e.g. angle, location or object count) is not invariant to these operations.

Thus, we disable random horizontal flipping as preprocessing for dSprites-orientation,  SmallNORB-azimuth and dSprites-location tasks. 
Random cropping preprocessing is disabled for Clevr-count,  Clevr-distance, DMLab, KITTI-distance and dSprites-location tasks.

\clearpage

\section{Robustness: Objects out-of-context}

\label{sec:random-context}

\begin{figure}[t]
\begin{center}
   \includegraphics[height=2.8cm]{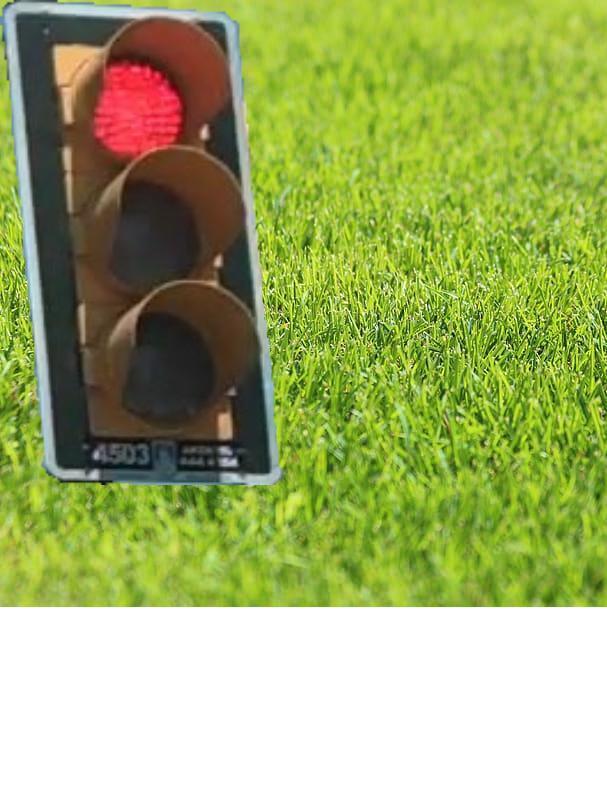}
   \raisebox{0.0cm}{\includegraphics[height=3.2cm]{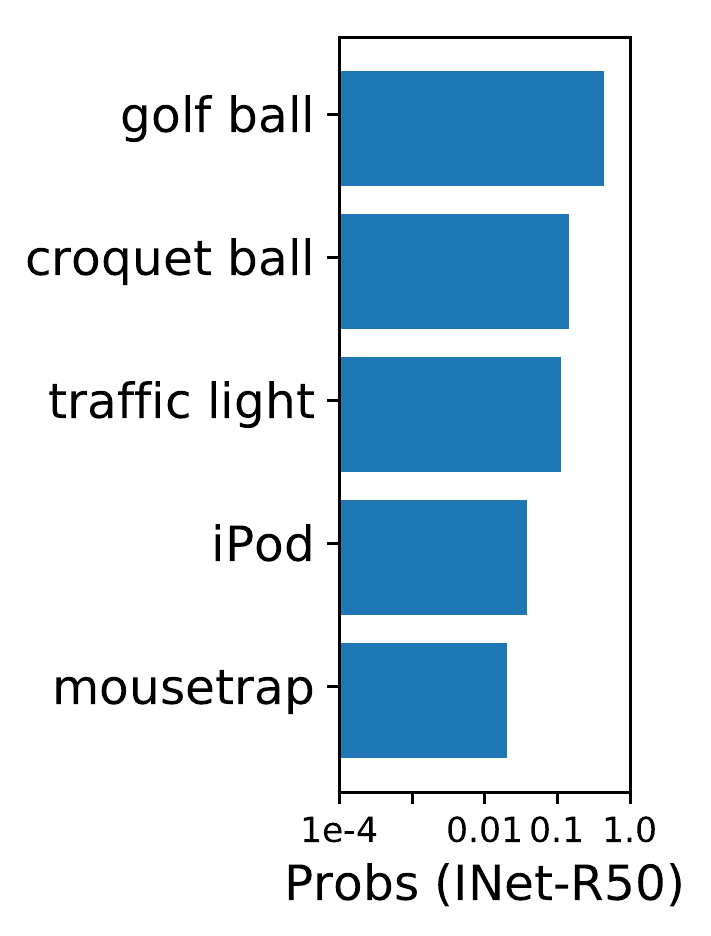}}
   \raisebox{0.0cm}{\includegraphics[height=3.2cm]{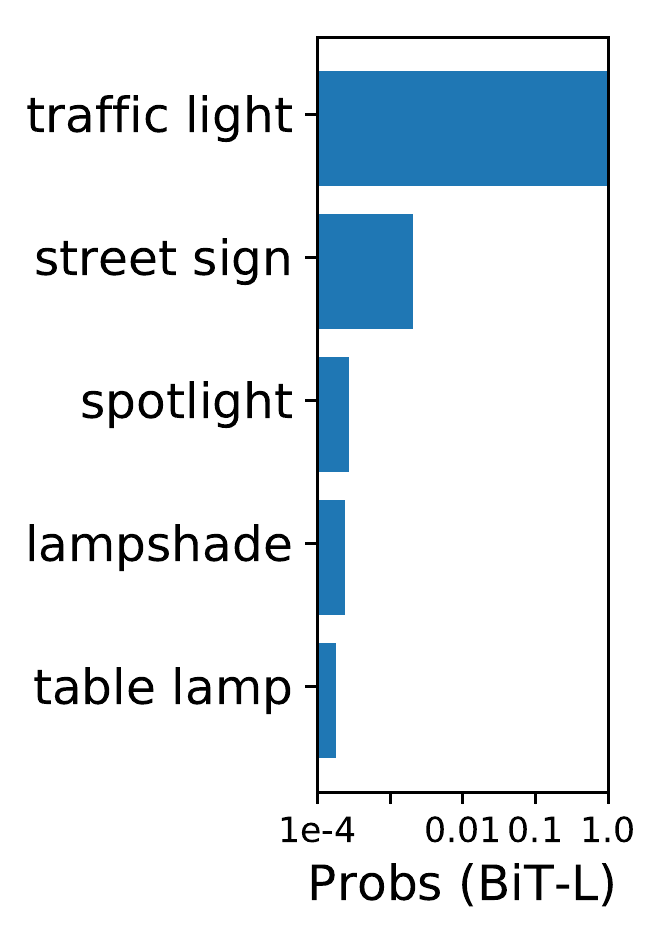}}
   \includegraphics[height=3.4cm]{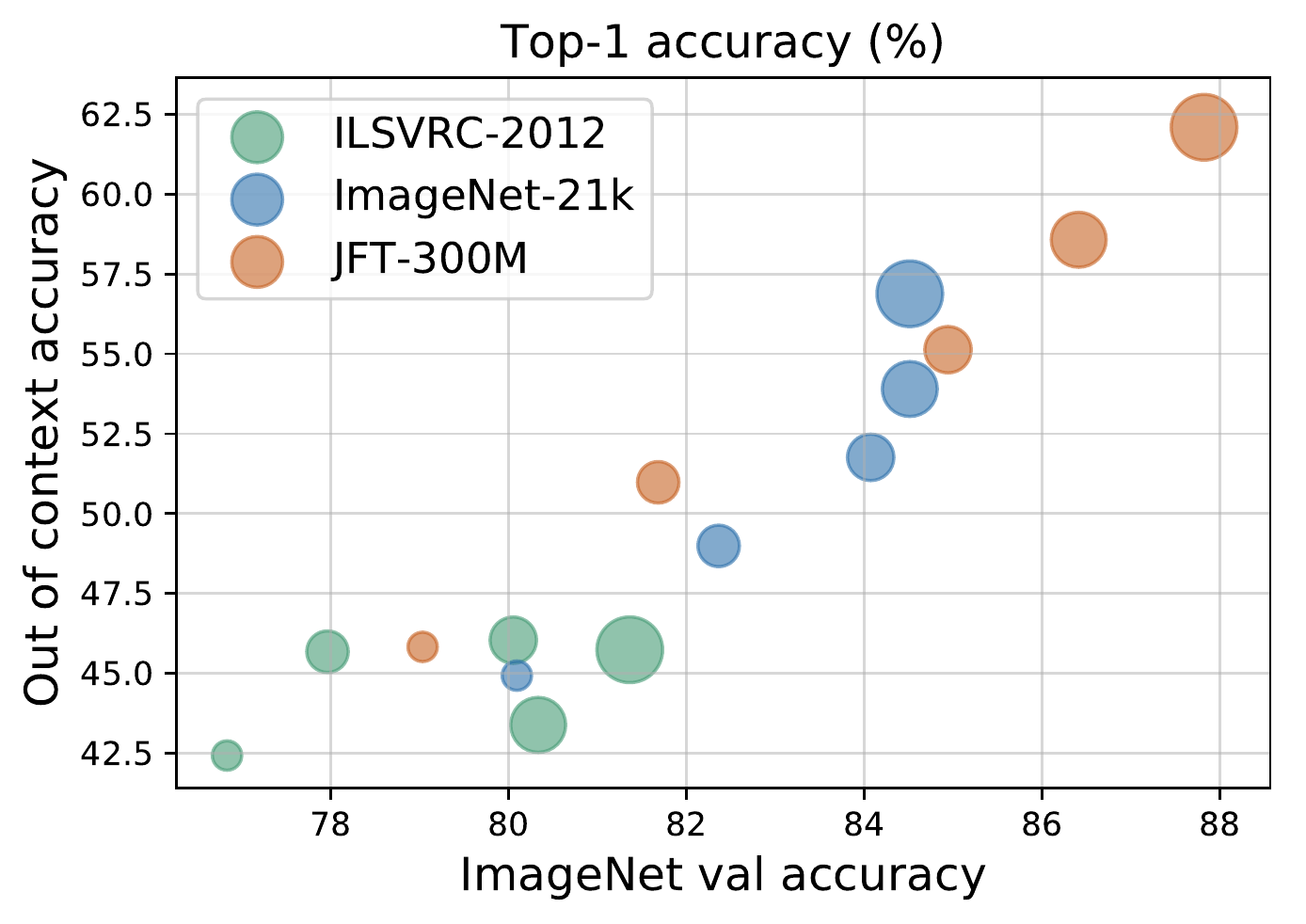}
\end{center}
\caption{
\textbf{Left:}
Top 5 predictions produced by an \imagenet{} model (IN-R50) and \name{}-L on an example out-of-context object. 
Bar lengths indicate predictive probability on a log scale.
\textbf{Right:}
Top-1 accuracy on the \imagenet{} validation set plotted against top-1 accuracy on objects out-of-context.
The legend indicates the pre-training data. All models are subsequently fine-tuned on \imagenet{} with \hyper{}.
Larger markers size indicates larger architectures, as in Fig.~\ref{fig:size_vs_acc}.} 
\label{fig:random-context}
\end{figure}

It has been shown that CNNs can lack robustness when classifying objects out-of-context~\cite{beery2018,peyre2017,shetty2018}. 
We investigate whether BiT not only improves classification accuracy, but also out-of-context robustness.
For this, we create a dataset of foreground objects corresponding to \imagenet{} classes pasted onto miscellaneous backgrounds (Fig.~\ref{fig:random-context} left). 
We obtain images of foreground objects using OpenImages-v5~\cite{OpenImages} segmentation masks.
Figure~\ref{fig:random-context} shows an example, and more are given in Figure~\ref{fig:out-of-context-examples-appendix}.
Sometime foreground objects are partially occluded, resulting in an additional challenge. 

We transfer \name{} models pre-trained on various datasets to \imagenet{} and see how they perform on this out-of-context dataset.
In Figure~\ref{fig:random-context} we can see that the performance of models pre-trained on \imagenet{} saturates on the out-of-context dataset, whereas by using more data during pre-training of larger models, better performance on \imagenet{} \emph{does} translate to better out-of-context performance.

More qualitatively, when we look at the predictions of the models on out-of-context data, we observe a tendency for \name{}-L to confidently classify the foreground object regardless of the context,
while \imagenet{} models also predict objects absent from the image, but that could plausibly appear with the background.
An example of this is shown in Figure~\ref{fig:random-context}~left.

\subsection{Out of context dataset details}
\label{out-of-context-dataset-details}
We generate this dataset by combining foreground objects extracted from OpenImages~V5~\cite{OpenImages} with backgrounds, licensed for reuse with modification, mined from search engine results. 

\textbf{Foreground objects} 
In this study, we evaluate models that output predictions over \imagenet{} classes.
We therefore fine-tune BiT models on \imagenet{} using \hyper{}.
We choose 20 classes from OpenImages that correspond to one such class or a subset thereof. These 20 classes cover a spectrum of different object types. We then extract foreground objects that belong to these classes from images in OpenImages using the provided segmentation masks.
Note that this leads to some objects being partially occluded; however, humans can still easily recognize the objects, and we would like the same from our models.

\textbf{Backgrounds} We define a list of 41 backgrounds that cover a range of contexts such that (1) we have reasonable diversity, and (2) the objects we choose would not likely be seen in some of these backgrounds. We then collect a few examples of each background using a search engine, limiting to results licensed for reuse with modification. We take the largest square crop of the background from the top left corner.

We paste the extracted foreground objects onto the backgrounds. This results in a total of 3321 images in our dataset (81 foreground objects $\times$ 41 backgrounds). We fix the size of the objects such that the longest side corresponds to 80\% of the width of the background; thus, the object is prominent in the image.

\begin{figure}[t]
\begin{center}
   \includegraphics[height=5cm]{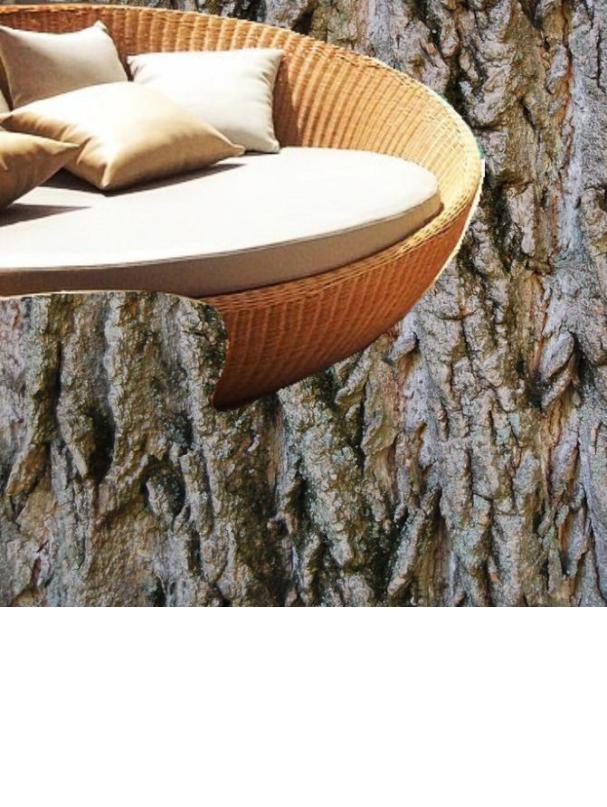}
   \raisebox{1.0cm}{\includegraphics[height=4cm]{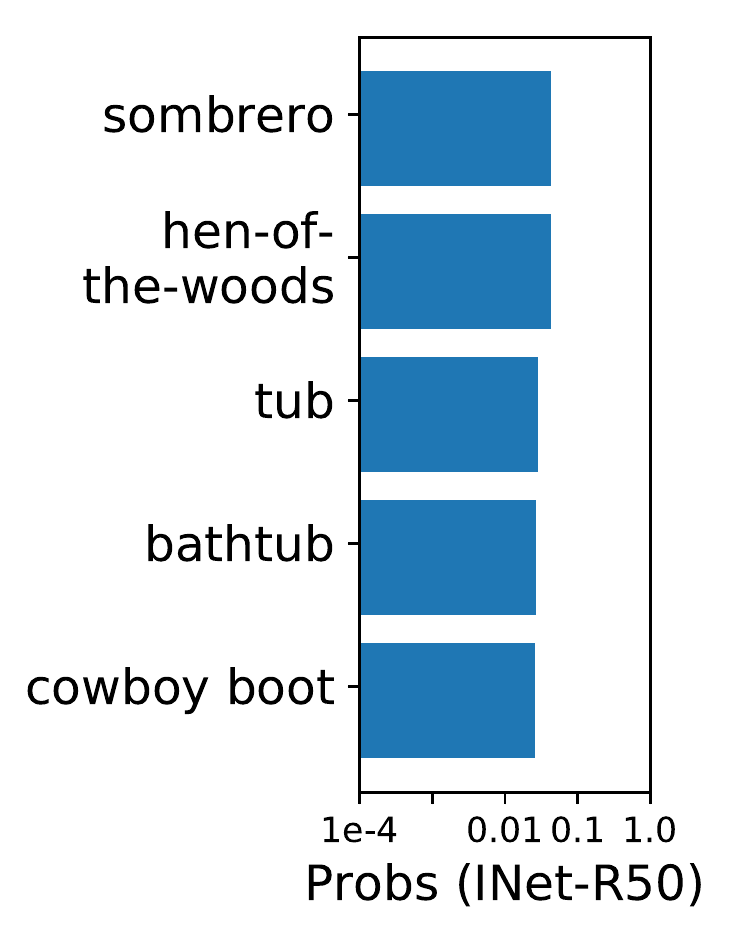}}
   \raisebox{1.0cm}{\includegraphics[height=4cm]{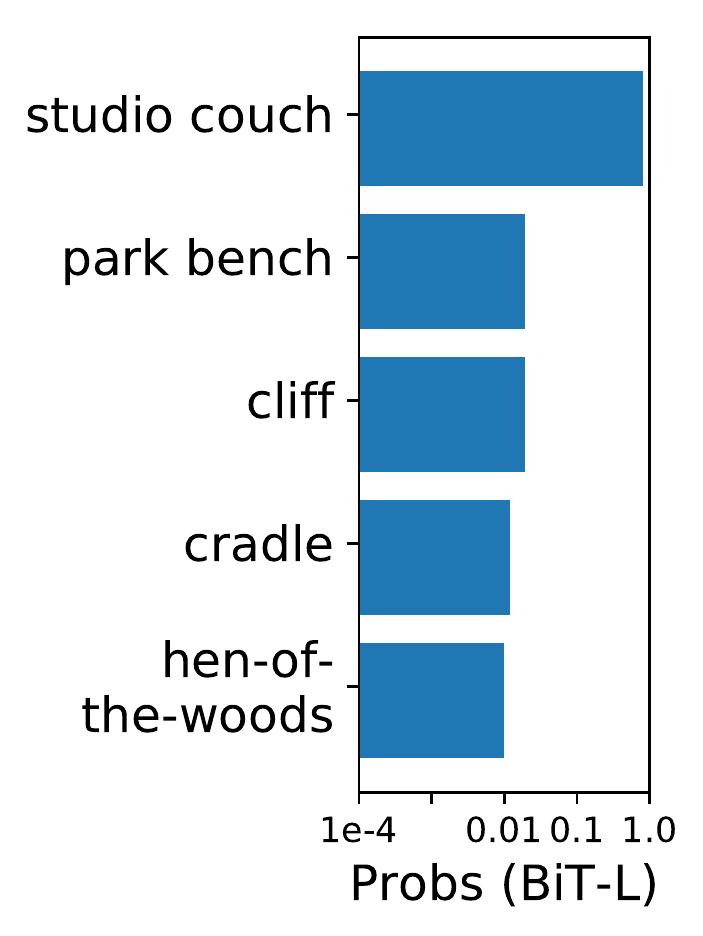}}
\end{center}\vspace{-40pt}%
\begin{center}
   \includegraphics[height=5cm]{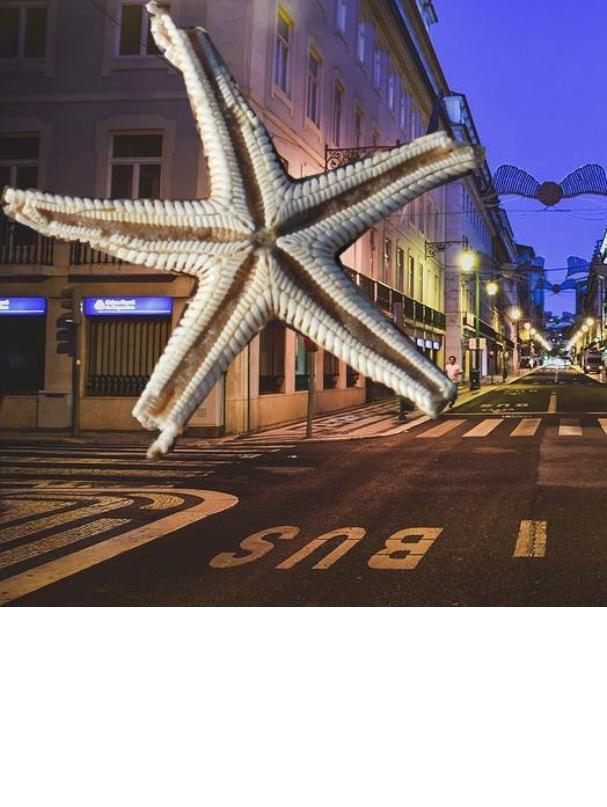}~
   \raisebox{1.0cm}{\includegraphics[height=4cm]{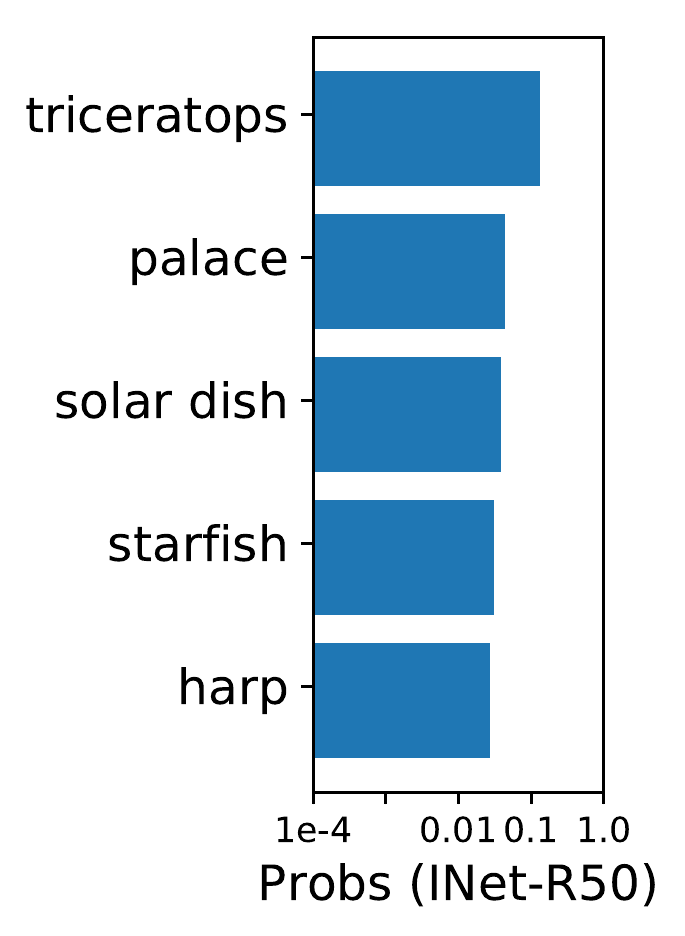}}
   \raisebox{1.0cm}{\includegraphics[height=4cm]{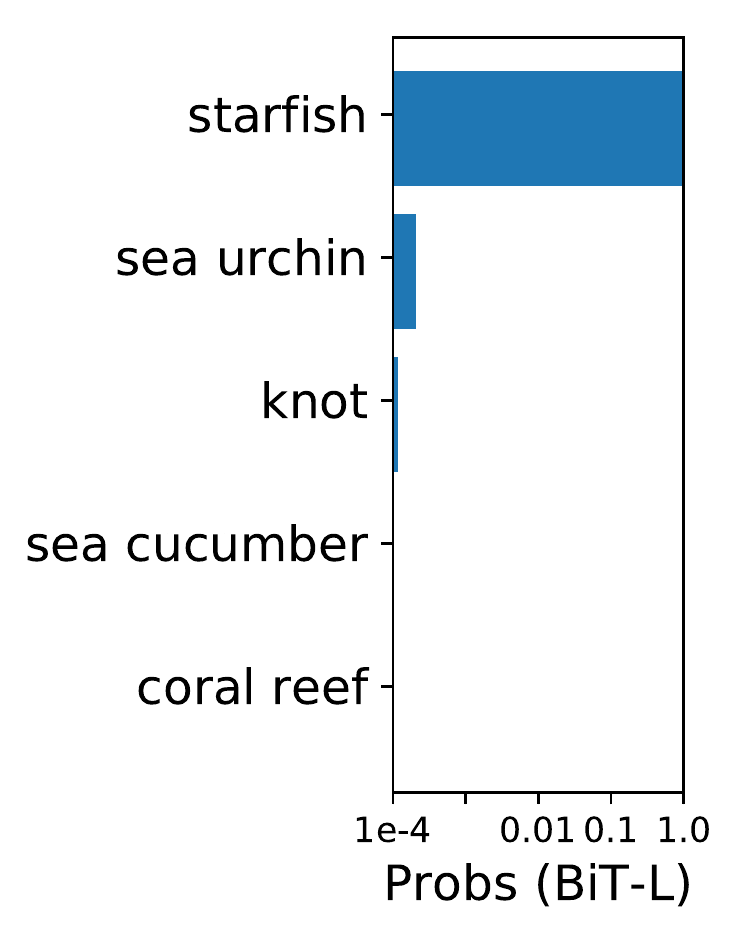}}
\end{center}\vspace{-40pt}%
\begin{center}
   \includegraphics[height=5cm]{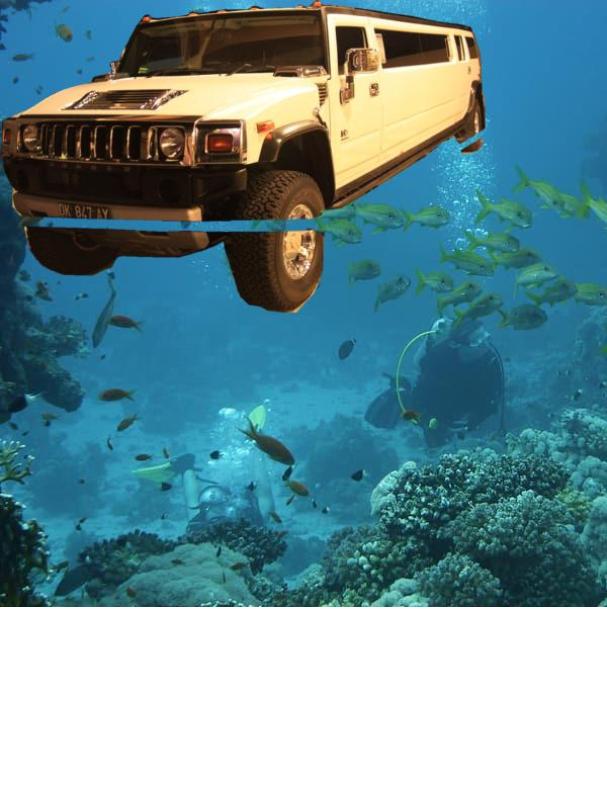}
   \raisebox{1.0cm}{\includegraphics[height=4cm]{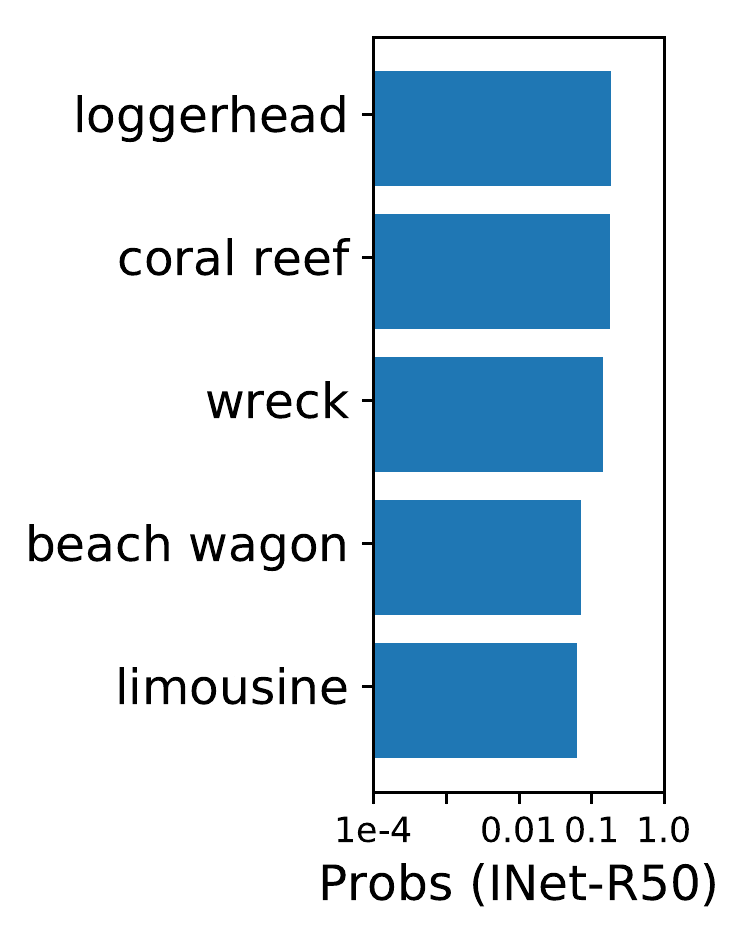}}
   \raisebox{1.0cm}{\includegraphics[height=4cm]{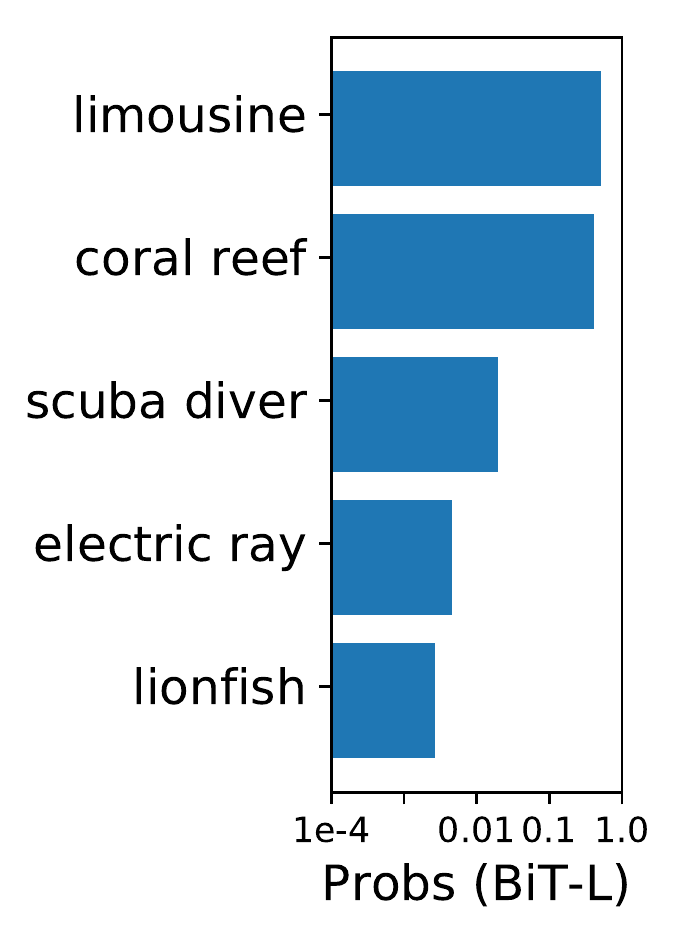}}
\end{center}\vspace{-40pt}%
\begin{center}
   \includegraphics[height=5cm]{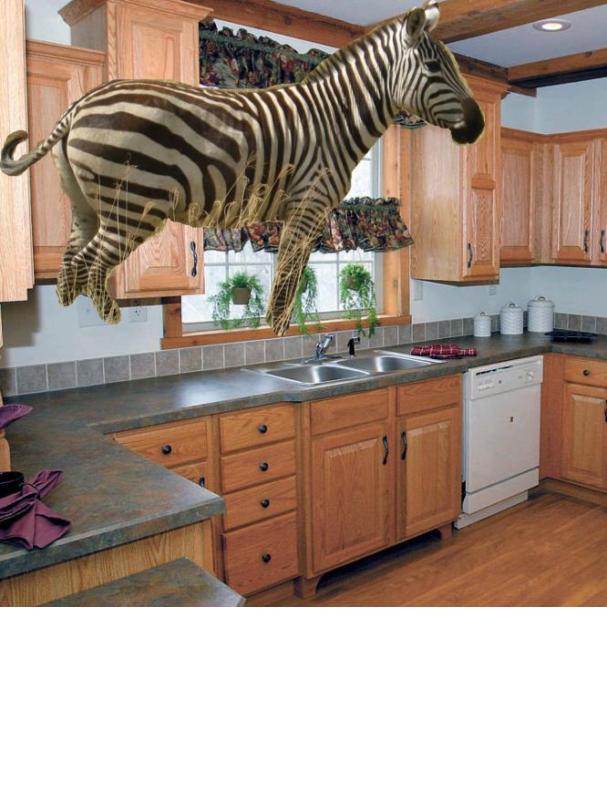}~
   \raisebox{1.0cm}{\includegraphics[height=4cm]{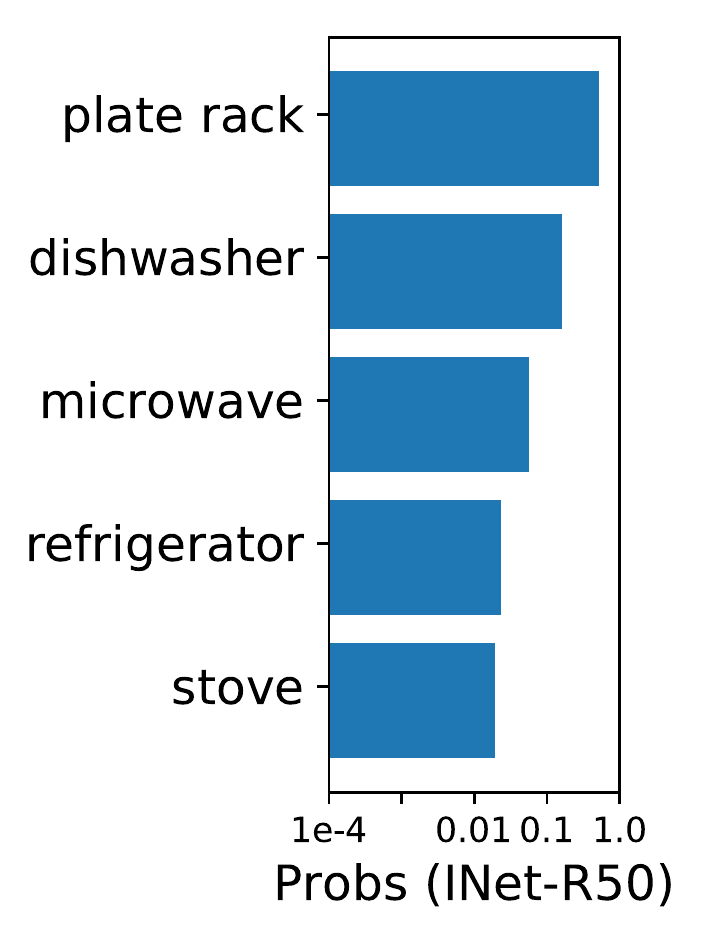}}
   \raisebox{1.0cm}{\includegraphics[height=4cm]{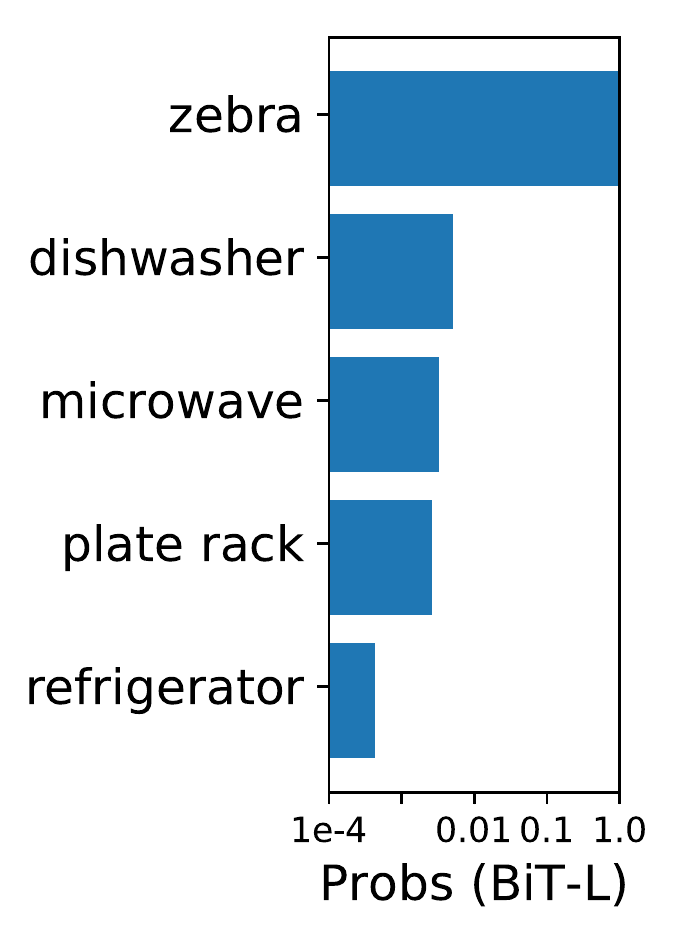}}
\end{center}\vspace{-40pt}%
\caption{Top 5 predictions produced by an \imagenet{} model (INet-R50) and \name{}-L on examples of out-of-context objects. 
Bar lengths indicate predicted probability on a log scale. We choose images that highlight the qualitative differences between INet-R50 and BiT-L predictions when the INet-R50 model makes mistakes.}
\label{fig:out-of-context-examples-appendix}
\end{figure}

Figure~\ref{fig:out-of-context-examples-appendix} shows more examples of out-of-context images from our dataset, contrasting the predictions given by a standard ResNet50 trained on \imagenet{} from scratch and the predictions of \name{}-L fine-tuned on \imagenet{}.

\subsection{Image Attributions}
In this section we provide attributions for images used to generate the examples from the out-of-context dataset.

\noindent All images are licensed CC-BY-2.0 unless noted otherwise.
\\ 

\noindent\textbf{Foreground objects:}
\begin{itemize}
   \item Traffic light: \href{https://farm3.staticflickr.com/3640/3576366805_eea15333d5_o.jpg}{U  turn to Tophane} by \emph{Istanbul Photo Guide}.
\item Sofa: \href{https://c8.staticflickr.com/4/3950/15557668485_5f23ed2ed3_o.jpg}{Welcome} by \emph{woot}.
\item Zebra: \href{https://farm8.staticflickr.com/3421/3367111780_dbfaf01821_o.jpg}{i like his tail in this one} by \emph{meg and rahul}.
\item Starfish: \href{https://c6.staticflickr.com/3/2936/14692943464_f618fc1360_o.jpg}{Starfish} by \emph{Summer Skyes 11}.
\item Limousine: \href{https://c7.staticflickr.com/5/4062/4271344866_1cb1a03cff_o.jpg}{Hummer limousine stopping at the door} [nb: title translated] by \emph{duncan\_su}. 
\end{itemize}

\noindent\textbf{Backgrounds:}
\begin{itemize}
    \item Grass: \href{https://www.pexels.com/photo/nature-field-grass-lawn-2637456/}{Photo} by \emph{zoosnow}\\
    (Pexels license; Free to use, no attribution required).
    \item Wood: \href{https://www.flickr.com/photos/31288116@N02/3752674533/sizes/l}{Tree Bark Texture 04} by \emph{Jacob Gube, SixRevisions}.
    \item Street at night: \href{https://pixabay.com/photos/city-street-calm-buildings-3875530/}{City street calm buildings} by \emph{csr\_ch}\\
    (Pixabay license; Free for commercial use, no attribution required).
    \item Underwater: \href{https://pixabay.com/ru/photos/\%D1\%80\%D0\%B0\%D0\%B7\%D0\%BB\%D0\%B8\%D1\%87\%D0\%BD\%D1\%8B\%D0\%B9-\%D1\%80\%D0\%B8\%D1\%84-\%D0\%B4\%D0\%B0\%D0\%B9\%D0\%B2\%D0\%B8\%D0\%BD\%D0\%B3-\%D0\%BA\%D0\%BE\%D1\%80\%D0\%B0\%D0\%BB\%D0\%BB-1198559/}{Photo} by \emph{MaxX42}\\ 
    (Pixabay license; Free for commercial use, no attribution required).
    \item Kitchen: \href{https://it.wikipedia.org/wiki/File:Modular_Kitchen.jpg}{Interior of a modern modular home} by \emph{Riverview Homes, Inc.}\\ 
    (CC-BY-SA-3.0 Unported license).
\end{itemize}